\documentclass[10pt,twocolumn,letterpaper]{article}

\usepackage{iccv}
\usepackage{times}
\usepackage{epsfig}
\usepackage{graphicx}
\usepackage{amsmath}
\usepackage{amssymb}

\usepackage[pagebackref=true,breaklinks=true,letterpaper=true,colorlinks,bookmarks=false]{hyperref}

\iccvfinalcopy 

\def\iccvPaperID{2509} 

\begin{document}

\title{Deep Single Image Deraining Via Estimating Transmission and Atmospheric Light in rainy Scenes}

\author{Yinglong Wang{$^{\dag}$}, Qinfeng Shi{$^{\ddag}$}, Ehsan Abbasnejad{$^{\ddag}$}, Chao Ma{$^{\S}$}, Xiaoping Ma{$^{\P}$}, Bing Zeng{$^{\dag}$}\\
$^{\dag}$School of Information and Communication Engineering,\\ University of Electronic Science and Technology of China\\
$^{\ddag}$School of Computer Science, The University of Adelaide\\
$^{\S}$Shanghai Jiao Tong University\\
$^{\P}$Beijing Jiao Tong University\\}

\maketitle

\begin{abstract}
   Rain removal in images/videos is still an important task in computer vision field and attracting attentions of
more and more people.
Traditional methods always utilize some incomplete priors or filters (\eg guided filter) to remove
rain effect. Deep learning gives more probabilities to better solve this task.
However, they remove rain either
by evaluating background from rainy image directly or learning a rain residual first then subtracting
the residual to obtain a clear background. No other models are used in deep learning based de-raining methods to
remove rain and obtain other information about rainy scenes.
In this paper, we utilize an extensively-used image degradation model which is
derived from atmospheric scattering principles to model the formation of rainy images and
try to learn the transmission, atmospheric light in rainy scenes and remove rain further.
To reach this goal, we propose a robust evaluation method of global atmospheric light in a rainy scene.
Instead of using the estimated atmospheric light directly to learn a network to calculate transmission,
we utilize it as ground truth and design a simple but novel triangle-shaped network structure to learn atmospheric
light for every rainy image, then fine-tune the network to obtain a better estimation
of atmospheric light during the training of transmission network.
Furthermore, more efficient ShuffleNet Units are utilized in transmission network to
learn transmission map and the de-raining image is then obtained by the image degradation model.
By subjective and objective comparisons, our method outperforms the selected state-of-the-art works.
\end{abstract}

\section{Introduction}
\begin{figure}[!t]
\begin{center}
\begin{minipage}{0.32\linewidth}
\centering{\includegraphics[width=1\linewidth]{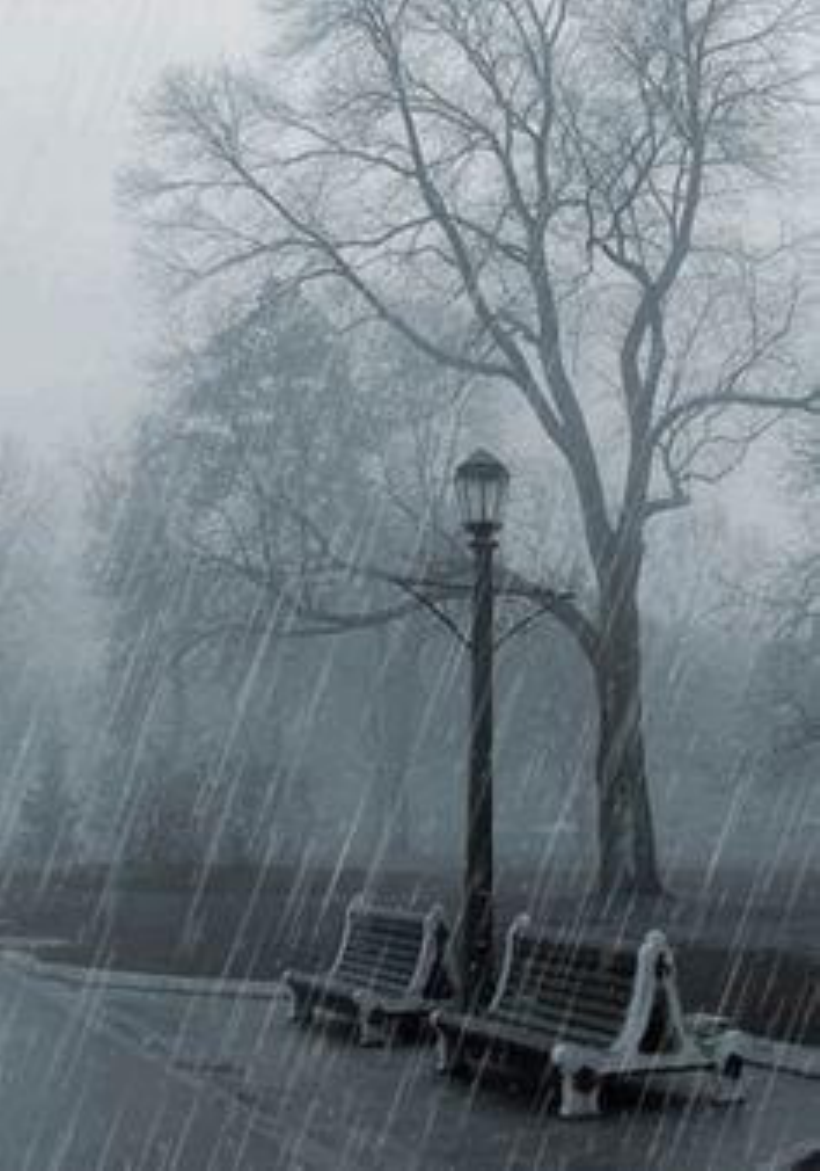}}
\centerline{Rainy image}
\end{minipage}
\begin{minipage}{0.32\linewidth}
\centering{\includegraphics[width=1\linewidth]{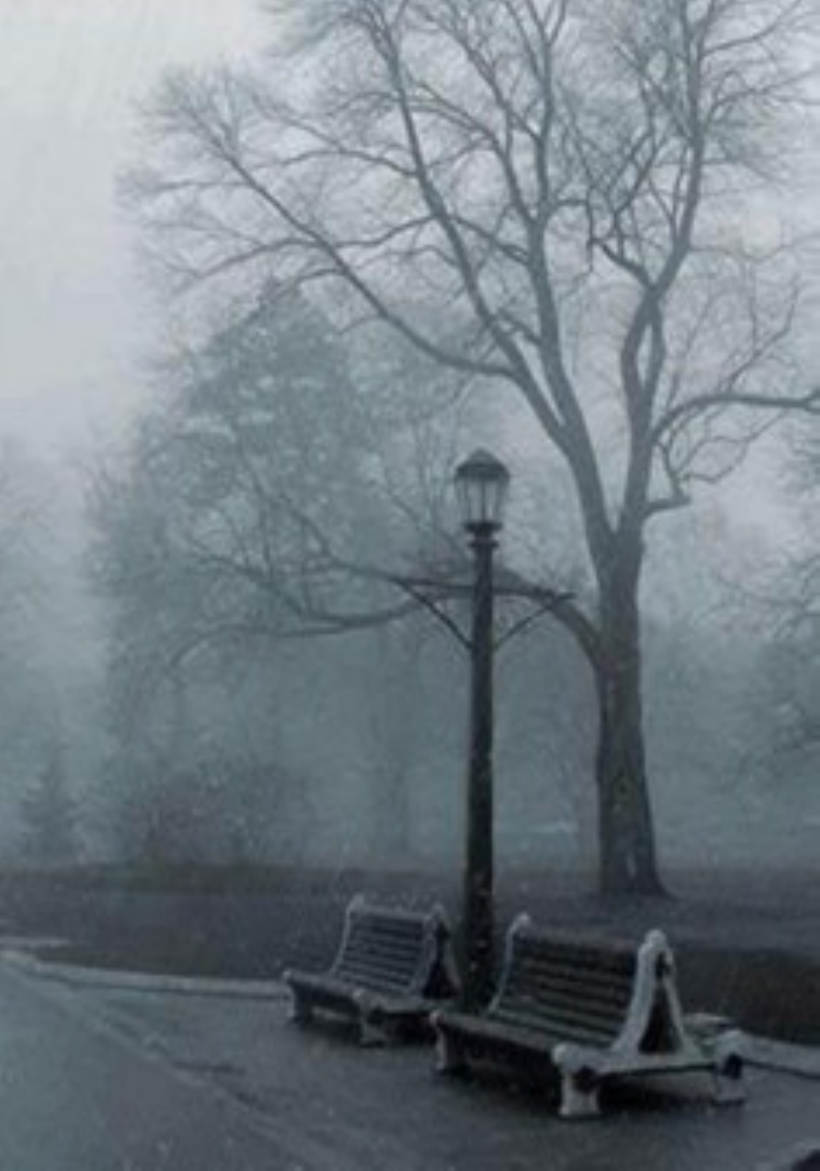}}
\centerline{De-raining result}
\end{minipage}
\begin{minipage}{0.32\linewidth}
\centering{\includegraphics[width=1\linewidth]{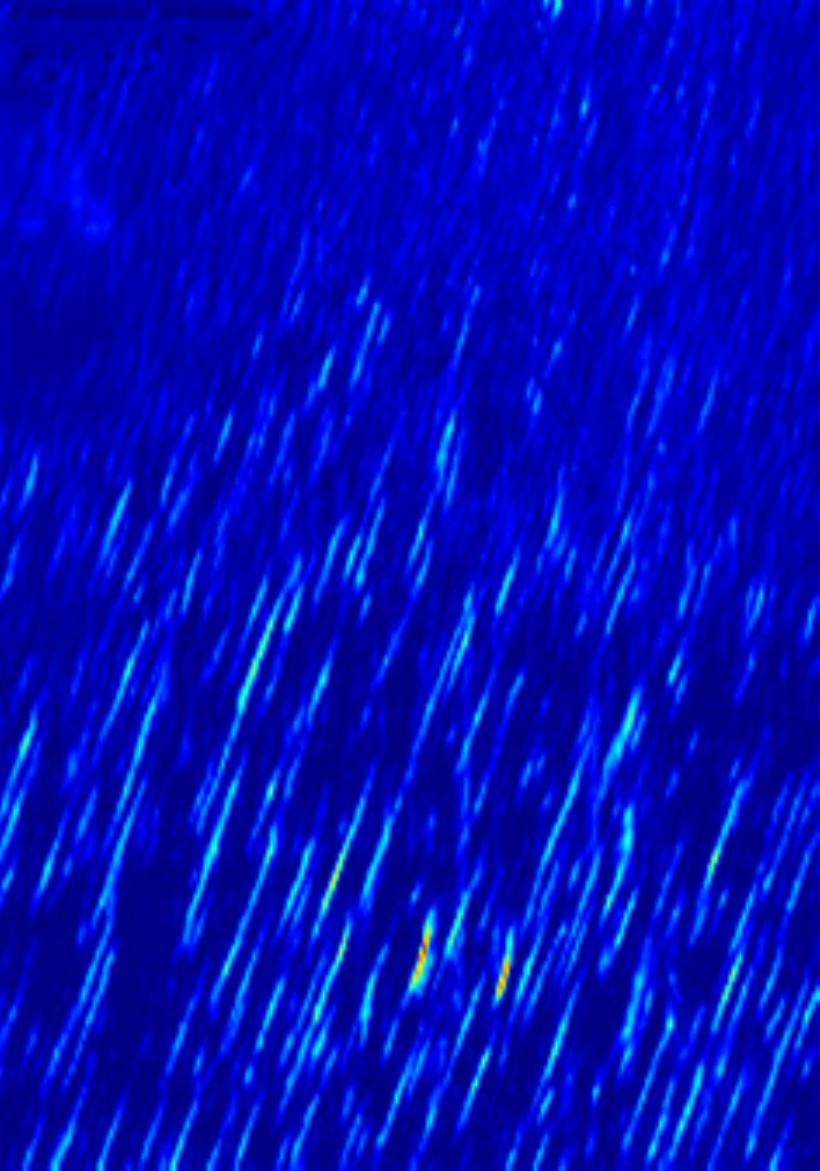}}
\centerline{Transmission}
\end{minipage}
\end{center}
\caption{An example of a real-world rainy image, its de-raining result and the transmission of rain. Our method obtains a clear rain-removed result, and the transmission of rain also gives us more details of rain, including the shapes, sizes, merged rain streaks and the degree of attenuation to the reflected light by background.}
\label{fig:first_example}
\end{figure}

Images and videos are becoming ubiquitous for keeping our favorite memories alive. However, the visual quality of outdoor images and videos are highly affected by weather conditions. For instance, rain occludes and distorts parts of the image that might be of great interest for both personal and commercial users. This, consequently, degrades the performances of many computer vision algorithms \eg \cite{Bahnsen_2018_TITS}. Typically rain and other distortions in images and videos are dealt with by \emph{denosing} where the objective is to remove additional undesired artifacts in the image.
In applications where outdoor images are primarily used (such as driverless cars\footnote{See the Bloomberg Businessweek article `Self-driving cars can handle neither rain nor sleet nor snow' on 17 Sept. 2018}), urgent attention to denoising is required. In particular, rain as one of the most prominent forms of artifacts in outdoor scenes poses unique set of challenges.
The pattern of rain streaks are complex, and some of them even merge into the background and cannot be seen. Some small image features which are utilized by some outdoor vision systems and computer vision algorithms will be destroyed seriously. Besides, due to the scattering of rain to light, the contrast of an image becomes really low, which also influences the performances of some algorithms.

Conventional rain removing methods in videos exploited the correlation information among frames to localize and remove rain streaks in either spatial or frequency domain, \eg \cite{Barnum_2007_PACV,Barnum_2010_CV,Bossu_2011_CV,Brewer_2008_CS}. For single images, however, rain removal is always more challenging as the problem is indeed an ill-posed one \cite{Zhang_2018_CVPR}.
There are no algorithms or models by which we can obtain accurate de-raining results. All the existing methods can only give an evaluation.
Dictionary-learning method \cite{Mairal_2010_JMLR} was first utilized to remove rain in single images \cite{Fu_2011_ASSP,Kang_2012_TIP,Chen_2014_CSVT,Wang_2016_ICIP,Huang_2014_TMM,Luo_2015_ICCV,Wang_2017_TIP}.
These approaches are computationally demanding when learning an over-complete dictionary. Besides, the feature descriptors that are utilized to identify rain atoms are not accurate, as such often image details are mistreated as rain and vice versa.
Gaussian mixture model-based layer priors \cite{Li_2016_CVPR} accommodate multi-scale and multi-orientation rain streaks, but occassionally miss image details and is computationally demanding. 

Deep learning has emerged as a powerful tool in achieving state-of-the-art results in many applications and showed to improve de-raining performances greatly. These improvements are not only in visual quality but also in processing speed \cite{Fu_2017_CVPR,Fu_2017_TIP,Yang_2017_CVPR,Zhang_2018_CVPR,Li_2018_MM}. These methods either (1) learn a clear image directly from the given rainy one or (2) estimate a rain residual which is subtracted from the observed rainy image to obtain the resulting de-rained one. However physical aspects of the scene such as atmospheric light and transmission are generally ignored.
These physical variables or an image degradation model which is derived from robust physical principle describe the influence of atmospheric particles (including rain) on images more completely and accurately. An accurate model also boost the de-raining performances. Besides, existing de-raining methods tend not to obtain good results for rain streaks whose edges are blurry and merge into the background.

To solve the above problems, we utilize an image degradation model which is derived from the principles of scattering medium (\eg fog, rain and snow) to describe the formation of rainy images.
This model reflects the interaction of light and medium and expresses the influence of bad weather on images more accurately \cite{Fattal_2008_TOG}.
However, this model was always used in the dehazing task but not de-raining. The major issue is that dynamic rain causes random spatial variations which are more hard to evaluate the atmospheric light and transmission than the static haze.

In this paper, we consider using the environment's physical parameters in a deep learning model for de-raining. In particular, we propose a novel neural architecture that is composed of two components that are jointly trained in an end-to-end fashion: (1) a transmission and (2) atmospheric light estimator. We show the combination of these two aspects learns to remove rain significantly better and achieves state-of-the-art results. In this framework, we first pre-train a triangle-shaped network with initial estimates of the atmospheric light. Subsequently, in joint training with the transmission, we fine-tune it. This approach allows us to handle wide rain streaks with blurred edges for which the existing methods performance is sub-par.
In Figure \ref{fig:first_example}, we show one example of rain-removed result from the given rainy image along with the transmission found by our model.  In the transmission, rain streaks become more apparent, especially these slim rain streaks which are merged in the background and cannot be seen in the original rainy image as well as the shape, size, and density. The whole map clearly shows the attenuation of rain to the reflected light.

In summary, the contributions of this paper are:
\begin{itemize}
\item
We utilize an image degradation model to describe the formation of rainy images, based on which, we calculate the atmospheric light, transmission, and de-raining result simultaneously in a rainy scene.
\item
We propose a method to evaluate the atmospheric light in a single rainy image, which has been an unsolved issue.
\item
We design a triangle-shaped network which utilizes the global image information to learn the global atmospheric light, and more efficient revised ShuffleNet units are utilized to construct our network to learn transmission in the rainy scenes.
\item
Compared to the state-of-the-art de-raining works, our method obtains better rain-removed images in objective and subjective assessments. We also show the generalization of our model by using it on de-hazing task.
\end{itemize}

\section{Related Work}

\noindent \textbf{Conventional Methods}
Rain effect was first removed in videos by utilizing the correlation information among video frames.
In this paper, we focus on single image de-raining task. Before 2010, learning an over-complete dictionary to express an image sparsely had had very good performance \cite{Mairal_2010_JMLR}. In a over-complete dictionary, every atom contains some specific image contents. An image can be expressed by these atoms sparsely. Fu \etal \cite{Fu_2011_ASSP} naturally used a dictionary to decompose a rainy image, so that some dictionary atoms just include image contents, while the other atoms only contain rain information.
Then they use the edge direction of rain to classified rain and non-rain atoms. The rain removed results then can be recovered by orthogonal matching pursuit algorithm \cite{Mallat_1993_TSP}. Some improved works which also use dictionary learning appeared afterward \cite{Kang_2012_TIP,Chen_2014_CSVT,Huang_2014_TMM,Luo_2015_ICCV,Wang_2016_ICIP,Wang_2017_TIP}. These works designed more robust feature descriptors to enhance the accuracy of identifying rain atoms. Different from other methods the work \cite{Wang_2017_TIP} analyze the common characteristics of rain and snow to design some descriptors (SVCC, PDIP) to remove both rain and snow. However, dictionary based methods are always time-consuming when training an over-complete dictionary. Another important shortcoming which influences the de-raining performance is the non-adaptive feature descriptors which tend not to completely express various rain, so that the generalization of these methods is always low.

To avoid time-consuming dictionary learning stage, some filter (\eg, the edge-preserving guided filter \cite{He_2013_PAMI}) based de-raining works appeared \cite{Ding_2015_MTA,Xu_2012_CIS}. These works are always simple, so they have high processing speed, but their de-raining effect is really limited. In \cite{Chen_2013_ICCV}, Chen \etal proposed a low-rank appearance model to capture the spatio-temporally correlated rain streaks. Li \etal \cite{Li_2016_CVPR} proposed priors which are based on Gaussian mixture models for both rain and background to accommodate multiple orientations and scales of the rain streaks. However, these two models will mistreat some image details as rain and remove them with rain streaks together. Besides, the work \cite{Li_2016_CVPR} is more time-consuming than dictionary learning methods.

\noindent \textbf{Deep Learning Based Methods} In recent three or four years, deep learning has been widely used in rain removal task and obtain better rain-removing performances compared with conventional methods. However, all these methods either estimate backgrounds directly from rainy images, or learn a rain residual layer which is subtracted from rainy image to obtain the rain-removed result. No other information about rainy scene is acquired.

Different from other common strategies, Fu \etal trained their DerainNet in high-frequency domain instead of image domain to extract image details to improve visual quality \cite{Fu_2017_TIP}.
In the meantime, inspired by deep residual network (ResNet) \cite{He_2015_CVPR}, a deep detail network which is also trained in high-pass domain was proposed to reduce the mapping range from input to output, to make the learning process easier \cite{Fu_2017_CVPR}. Though training networks on high-frequency details can remove the interference from the background to a degree, some bright rain streaks cannot be filtered completely from low-frequency part so that some rain streaks will remain in the final results for some rainy images.
Yang \etal decomposed a rainy image into rain layer and background layer and added a binary map to locate rain streaks. Besides, they created a new model which includes two components to express rain streak accumulation and various shape and directions of rain streaks \cite{Yang_2017_CVPR}. With the help of binary map, this method can deal bright rain streaks, but tend to neglect some rain streaks with blurring edges. In \cite{Zhang_2018_CVPR}, Zhang \etal tried to automatically estimate rain density by network itself, then a multi-stream densely connected DID-MDN structure which can better characterize rain streaks with various shape and size is trained to remove rain streaks guided by the estimated rain density. However, this work can cause blur for some images with fine details.

A multi-stage network which consists of several parallel sub-networks was designed to model and remove rain streaks of various size \cite{Li_2017_arxiv}. Different parallel sub-networks model rain streaks with corresponding sizes.
Li \etal regarded rain streaks as the accumulation of multiple rain streaks layers, then use a recurrent neural network to remove rain streaks state-wisely. Though, recurrent training method is used, this work is not sensitive to rain streaks with blur edges.
In \cite{Li_2018_MM}, a non-locally enhanced encoder-decoder network framework is proposed to capture long-range spatial dependencies via skip-connections and pooling indices guided decoding is used to learn increasingly abstract feature representation to preserve the image details decoding. Different from all these state-of-the-art de-raining works, we design a simple but effective network to estimate atmospheric light, transmission of rain and obtain a clear rain-removed results. For the wide rain streaks with blurry edges, our method also produces better visual effect.

\section{Background}

By 2001, Nayar and Narasimhan had made detailed studies to the influence of bad weather (\eg, haze, rain, snow) on images/videos in computer vision by utilizing the scattering and absorption principle of medium to light \cite{Nayar_1999_ICCV,Narasimhan_2002_IJCV}, and several expressions had be made to reflect the degradation of bad weather to images/videos. Global atmospheric light  can be approximated as a constant $3 \times 1$ vector, then a simplified image degradation model can be obtained \cite{Fattal_2008_TOG}:
\begin{equation}\label{eq:degradation_model}
\mathbf{I(\mathbf{x})} = \mathbf{T}(\mathbf{x})\mathbf{J}(\mathbf{x}) + (1-\mathbf{T}(\mathbf{x}))\mathbf{A},
\end{equation}
where $\mathbf{I(\mathbf{x})}$ is the observed intensity, $\mathbf{J(\mathbf{x})}$ is scene radiance \cite{He_2011_PAMI}, $\mathbf{A}$ is the global atmospheric light and $\mathbf{T}(\mathbf{x})$ is the medium transmission to describe the attenuation of medium to light.
$\mathbf{T}(\mathbf{x})\mathbf{J}(\mathbf{x})$ is the attenuated scene intensity (radiance) by the medium called direct attenuation \cite{Tan_2008_CVPR}. $(1-\mathbf{T}(\mathbf{x}))\mathbf{A}$ is the scattered light called airlight \cite{Koschmieder_1924_BPFA,Tan_2008_CVPR} which leads to the color shift of the scene.
The more general expression of transmission is:
\begin{equation}\label{eq:transmission}
\mathbf{T}(\mathbf{x})=e^{- \int_{0}^{d} \beta(\mathbf{x},s)ds},
\end{equation}
where $\beta(\mathbf{x},s)$ is the scattering coefficient of the medium \cite{He_2011_PAMI} which is relative to the pixel location $\mathbf{x}$ ($\mathbf{x}$ corresponds to a scene point) and the distance $d$ of scene to the camera.

\begin{figure}[!t]
\begin{center}
\begin{minipage}{0.19\linewidth}
\centering{\includegraphics[width=1\linewidth]{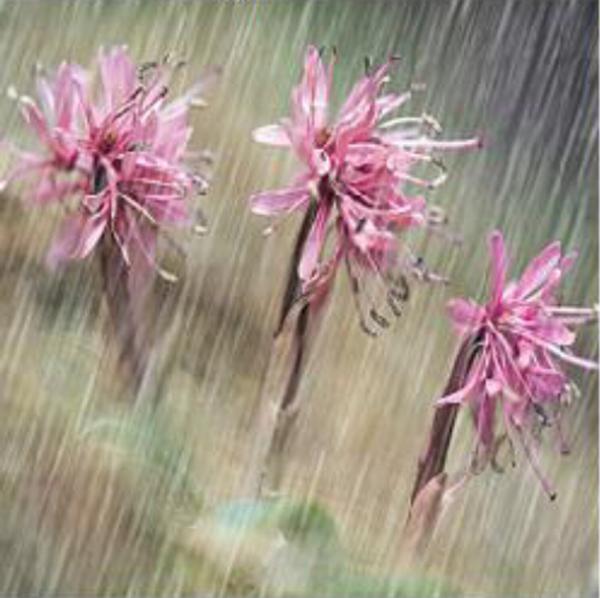}}
\centerline{(a)}
\end{minipage}
\hfill
\begin{minipage}{0.19\linewidth}
\centering{\includegraphics[width=1\linewidth]{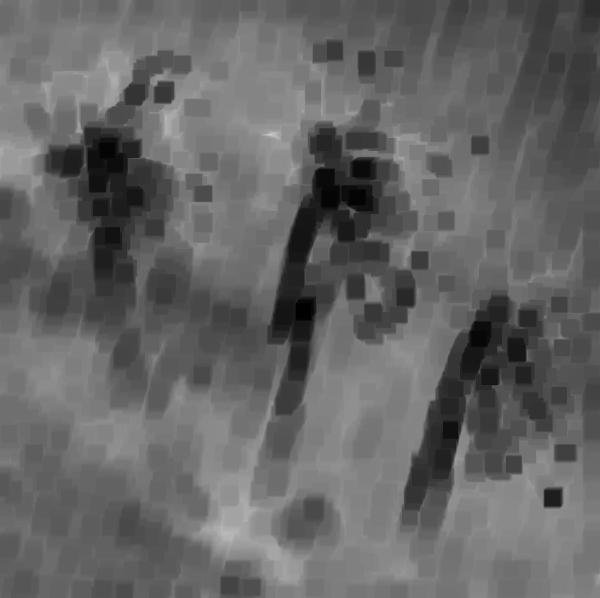}}
\centerline{(b)}
\end{minipage}
\hfill
\begin{minipage}{0.19\linewidth}
\centering{\includegraphics[width=1\linewidth]{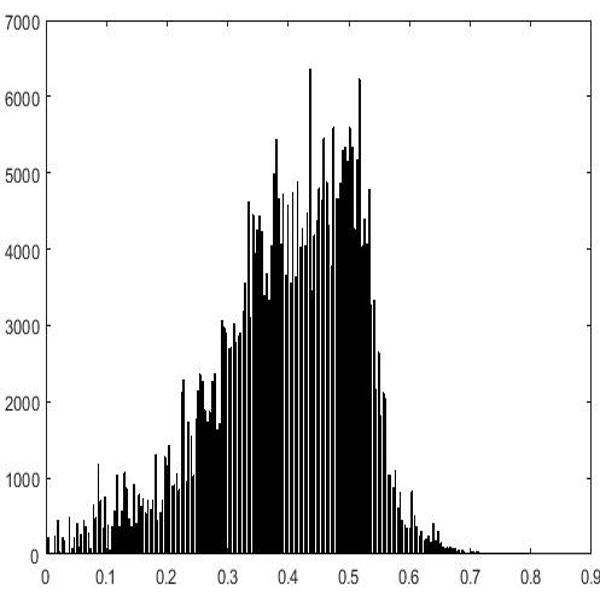}}
\centerline{(c)}
\end{minipage}
\hfill
\begin{minipage}{0.19\linewidth}
\centering{\includegraphics[width=1\linewidth]{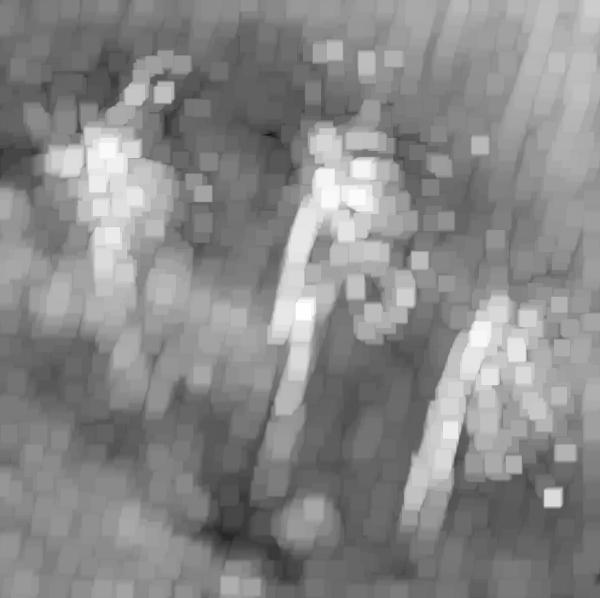}}
\centerline{(d)}
\end{minipage}
\hfill
\begin{minipage}{0.19\linewidth}
\centering{\includegraphics[width=1\linewidth]{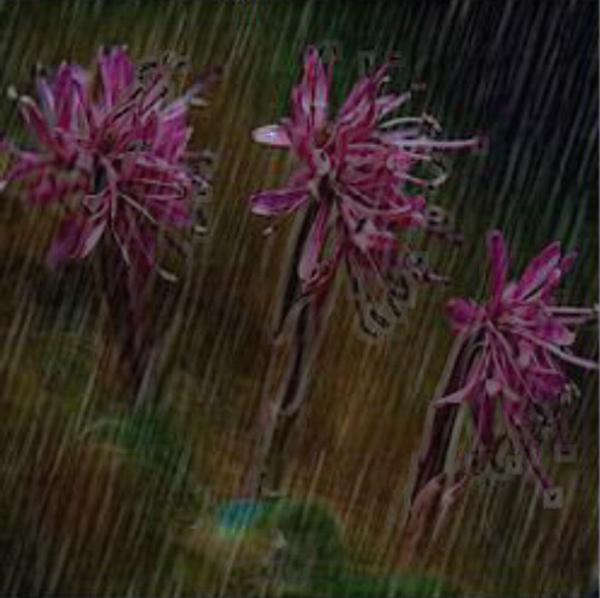}}
\centerline{(e)}
\end{minipage}
\end{center}
\caption{Some results of using dark channel prior to a rainy image. (a) Observed rainy image. (b) Dark channel prior. (c) Histogram of Dark channel prior. (d) Transmission estimated by dark channel prior. (e) De-raining result by dark channel prior.}
\label{fig:dark_channel}
\end{figure}

Haze is homogeneous approximately in a small patch, its scattering coefficient $\beta(\mathbf{x},s)$ can be regarded as a constant value, which gives great convenience to remove haze from single images by Eq. \ref{eq:degradation_model} \cite{Narasimhan_2002_IJCV,Nayar_1999_ICCV,Tan_2008_CVPR,Fattal_2008_TOG}. The dark channel prior \cite{He_2011_PAMI} helps to better evaluate transmission $\mathbf{T}(\mathbf{x})$ and the global atmospheric light $\mathbf{A}$. However, the assumption that $\beta(\mathbf{x}, s)$ is constant is not satisfied any more in rainy scenes. Hence, the following estimation of transmission \cite{He_2011_PAMI} cannot be obtained:
\begin{equation}\label{eq:estimate_transmission}
\widetilde{\mathbf{T}}(\mathbf{x}) = 1 - min_{c}(min_{\mathbf{y} \in \Omega(\mathbf{x})}(\frac{I^{c}(\mathbf{y})}{A^{c}})).
\end{equation}
Besides, the definition of dark channel:
\begin{equation}\label{eq:dark_channel}
J^{dark}(\mathbf{x}) = min_{c \in \{r,g,b\}} (min_{\mathbf{y} \in \Omega(\mathbf{x})} (J^{c}(\mathbf{y})))
\end{equation}
is also not significant, because rain streaks are merged into the dark channel, and some even disappear.
In the above two equations, $J^{c}$ is a color channel of $\mathbf{J}$, $\Omega(\mathbf{x})$ is a local patch centered at $\mathbf{x}$.
In Figure \ref{fig:dark_channel}, we show the transmission and dark channel of a rainy image calculated by \eqref{eq:estimate_transmission}\eqref{eq:dark_channel} respectively. From Figure \ref{fig:dark_channel} (b)(d), we can see that there is not so much rainy information in the dark channel and estimated transmission. Hence, the dark channel prior cannot estimate global atmospheric light $\mathbf{A}$ as like in \cite{He_2011_PAMI}. Figure \ref{fig:dark_channel}(e) is the de-raining result by the dark channel prior, which proves further that dark channel prior cannot be used in de-raining task.

Though Eq. \eqref{eq:degradation_model} is derived to describe the image degradation caused by the scattering of several kinds of atmospheric particles (\eg, haze, rain, snow), no one has used it in de-raining task before. In the next section, we will present a new method to estimate global atmospheric light $\mathbf{A}$ in rainy scenes, design a simple and efficient network to estimate transmission of rain and remove rain by Eq. \eqref{eq:degradation_model} further.

\section{Our Method}

\begin{figure*}[!t]
\begin{center}
\begin{minipage}{0.8\linewidth}
\centering{\includegraphics[width=1\linewidth]{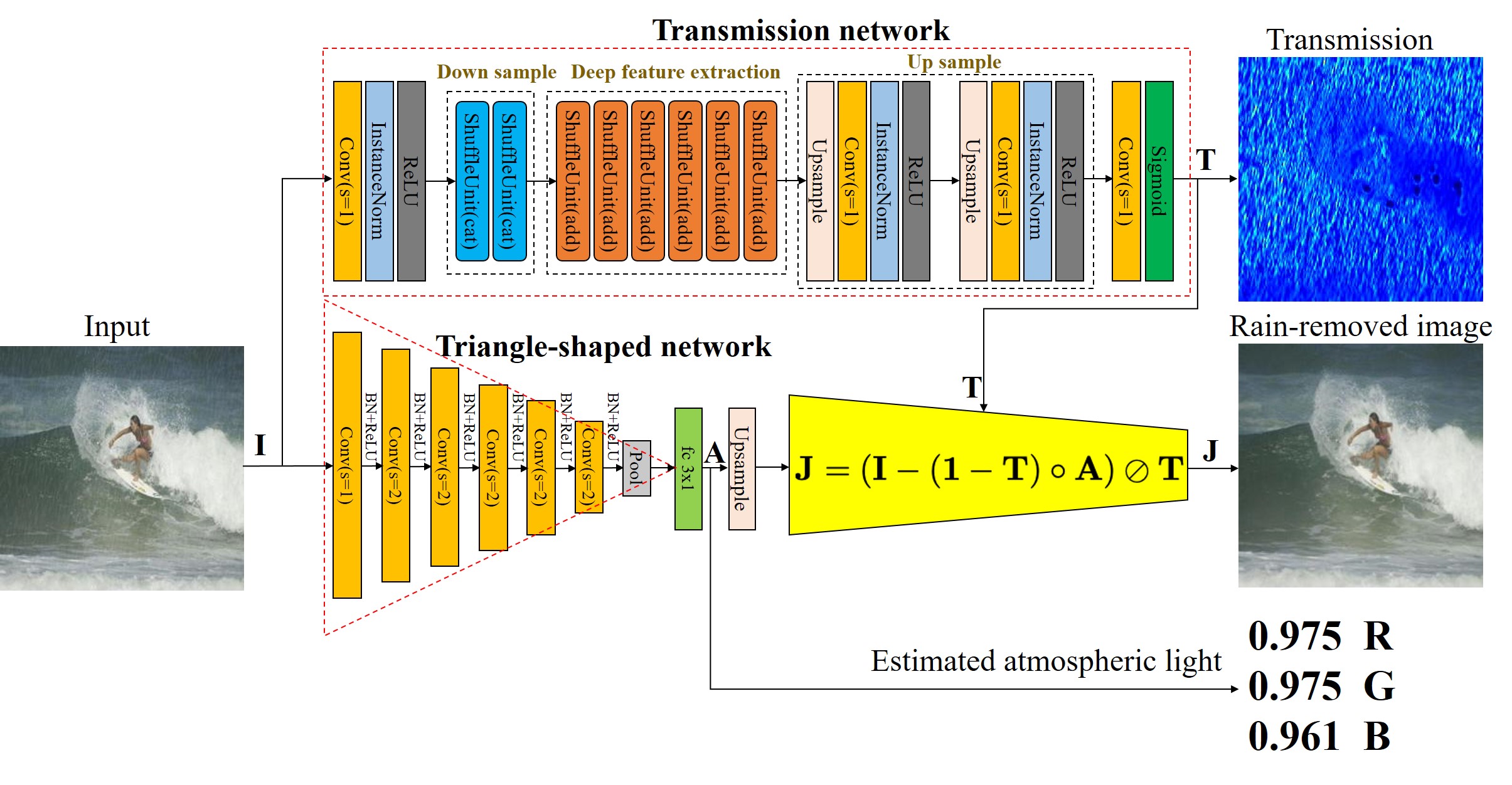}}
\end{minipage}
\end{center}
\caption{{This figure shows our whole network structure. $\circ$ and $\oslash$ are pixel-wise multiplication and division respectively. The 'Pool' operation means adaptive average pooling. The Upsample operation after triangle-shaped network just extends the atmospheric light to the image size.}}
\label{fig:whole_net}
\end{figure*}

Given a rainy image $\mathbf{I}$, the goal of our paper is to recover a rain-free image $\mathbf{J}$, estimate the global atmospheric light $\mathbf{A}$ and calculate rain's transmission $\mathbf{T}$. Our immediate thought is to build a DEgradation MOdel based Network (DEMO-Net) to estimate $\mathbf{J}$, $\mathbf{A}$ and $\mathbf{T}$ from a single rainy image $\mathbf{I}$:
\begin{equation}\label{eq:demo_model}
\mathbf{J}, \mathbf{A}, \mathbf{T} = \mathcal{D}(\mathbf{I}),
\end{equation}
where $\mathcal{D}(\cdot)$ denotes the mapping of our DEMO-Net and its whole network structure is shown in Figure \ref{fig:whole_net}. Different from many previous deep learning based works which learn $\mathbf{J}$ directly from $\mathbf{I}$ or estimate a rain residual $\mathbf{R}$ ($\mathbf{J} = \mathbf{I}-\mathbf{R}$). We utilize a more accurate and complete model to express rainy images to not only remove rain but also obtain two important physical variables in rainy scenes. Before introducing our network $\mathcal{D}(\cdot)$, we first estimate global atmospheric light $\mathbf{A}$ which is really important for our whole work.

\begin{figure}[!t]
\begin{center}
\begin{minipage}{0.85\linewidth}
\centering{\includegraphics[width=1\linewidth]{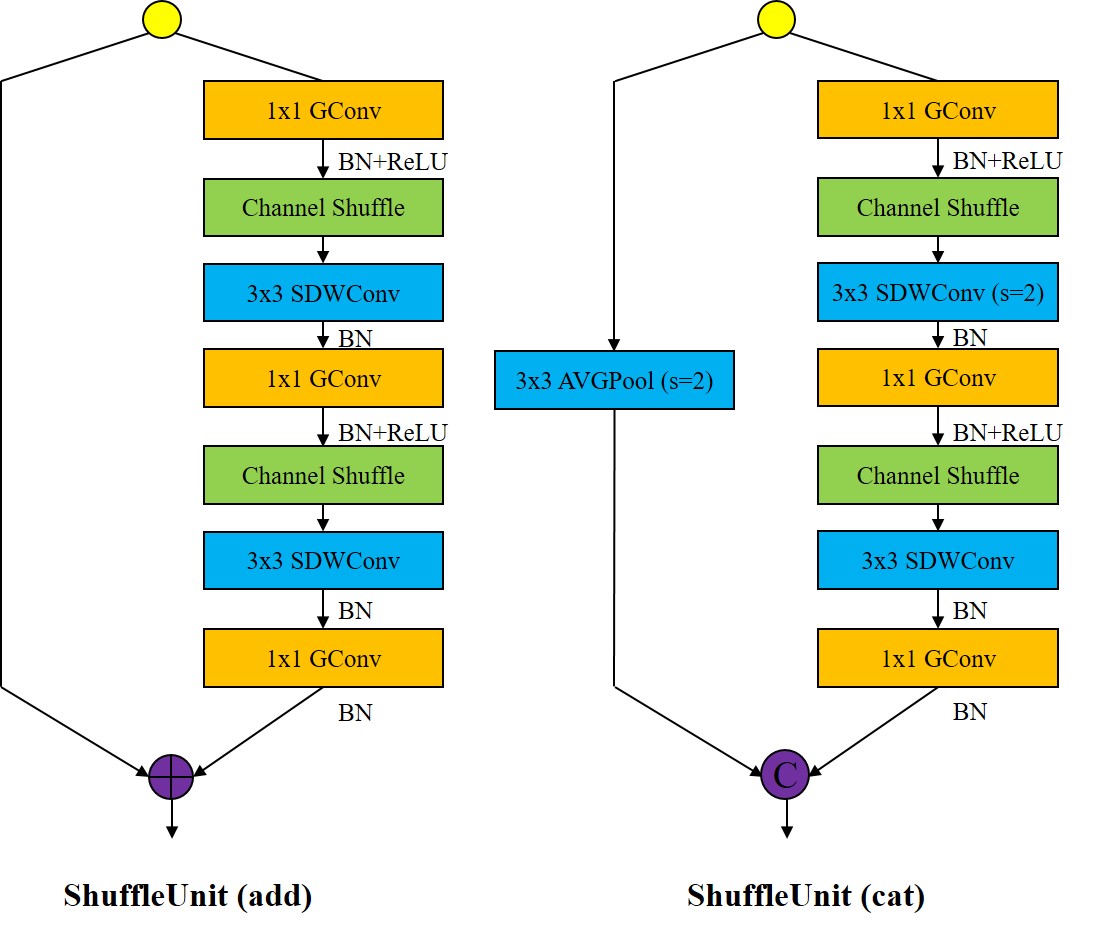}}
\end{minipage}
\end{center}
\caption{{This figure shows our revised ShuffleNet Units. ShuffleUnit(add) can keep image size unchanged, and ShuffleUnit(cat) downsamples image once. $+$ and C mean addition and concatenation respectively.}}
\label{fig:shuffle_unit}
\end{figure}

\subsection{Estimating the Global Atmospheric Light}
\label{sec:atmospheric_light}

Airlight ($(1-\mathbf{T}(x))\mathbf{A}$) is the result of $\mathbf{A}$ scattered by atmospheric particles \cite{Koschmieder_1924_BPFA}.
Hence, $\mathbf{A}$ can be approximated by the highest intensity of particles in images. Based on the same principle, a pixel with the highest intensity is used to estimate $\mathbf{A}$ for hazy images in \cite{Tan_2008_CVPR}, \cite{Fattal_2008_TOG,He_2011_PAMI} refine its estimation. In actual, He \etal \cite{He_2011_PAMI} utilized the dark channel prior to locate haze (the parts which have relatively high values in dark channel prior), then use the top $0.1\%$ brightest pixels in dark channel to determine $\mathbf{A}$. The influence of rain on images is more complex than haze and the dark channel prior loses function for rainy images.

According to \cite{Koschmieder_1924_BPFA}, in rainy scenes, the highest value of rain pixels can be used to estimate $\mathbf{A}$. So we must locate rainy pixels first, then find the highest value as the estimation of $\mathbf{A}$. In \cite{Yang_2017_CVPR}, Yang \etal designed a deep neural network to localise rainy pixels.
However, the authors used Softmax to approximate binary training samples. Hence, we use a threshold to obtain a binary location map, then find the rainy pixel with highest value to approximate $\mathbf{A}$. Some location maps are shown in supplement.

\subsection{Proposed DEMO-Net}

According to the image degradation model in Eq. \eqref{eq:degradation_model}, given a rainy image $\mathbf{I}$, we can obtain its rain-removed image by the following expression if the atmospheric light $\mathbf{A}$ and the transmission $\mathbf{T}$ can be known first:
\begin{equation}\label{eq:derain}
    \widehat{\mathbf{J}} = (\mathbf{I} - (\mathbf{1}-\mathbf{T})\circ\mathbf{A}) \oslash {\mathbf{T}},
\end{equation}
where $\widehat{\mathbf{J}}$ is the estimation of $\mathbf{J}$, $\circ$ and $\oslash$ denote pixel-wise multiplication and division respectively.

In Section \ref{sec:atmospheric_light}, we have obtained an initial evaluation for $\mathbf{A}$. In order to acquire an estimation of $\mathbf{A}$ which is suit better to our DEMO-Net to obtain a clear rain-removed result and good transmission, we train a triangle-shaped neural network to learn atmospheric light for every rainy image which is supervised by our estimated $\mathbf{A}$ in Section \ref{sec:atmospheric_light}. We use $\mathcal{A}(\cdot)$ to denote the mapping of this network. Training a network can synthesize the information from many scenes and obtain more robust estimation, which is equal to learning a general distribution of atmospheric light $\mathbf{A}$.

After obtaining $\mathbf{A}$, we utilize a novel network denoted as $\mathcal{T}(\cdot)$ to calculate the transmission $\mathbf{T}$. $\mathcal{T}(\cdot)$ and $\mathcal{A}(\cdot)$ are composed based on the model \eqref{eq:degradation_model}. By Eq. \eqref{eq:derain}, we finally calculate the rain-removed result (shown in Figure \ref{fig:whole_net}).
During the training for transmission, $\mathcal{T(\cdot)}$ and pre-trained $\mathcal{A(\cdot)}$ estimate $\mathbf{T}$ and $\mathbf{A}$ respectively, then Eq. \eqref{eq:derain} is used to calculate rain-removed result in the forward propagation.
Eq. \eqref{eq:derain} is differentiable to $\mathbf{A}$ and $\mathbf{T}$, hence in the back propagation, we use the difference between the ground-truth $\mathbf{J}$ and calculated de-raining result $\widehat{\mathbf{J}}$ to not only update the parameters of $\mathcal{T(\cdot)}$, but also fine-tune the parameters of $\mathcal{A(\cdot)}$ with smaller learning rate to boost the performances. This is also one of the reasons why we use network to re-estimate $\mathbf{A}$ rather than using initially-estimated $\mathbf{A}$ directly.

\subsection{Network Structure for $\mathcal{A(\cdot)}$}

$\mathcal{A(\cdot)}$ takes the rainy image $\mathbf{I}$ as input to estimate $\mathbf{A}$:
\begin{equation}\label{eq:mapping_A}
    \widehat{\mathbf{A}} = \mathcal{A}(\mathbf{\mathbf{I}}),
\end{equation}
where $\widehat{\mathbf{A}}$ is the estimation of $\mathbf{A}$. According to \cite{Koschmieder_1924_BPFA,Narasimhan_2002_IJCV,Nayar_1999_ICCV}, atmospheric light is a $3 \times 1$ vector. Hence, we design layer-wise down-sampling triangle-shaped convolution network, shown in Figure \ref{fig:whole_net}, so that we can obtain a $3 \times 1$ vector. A convolution layer is first used to extract features primarily. Then five convolution layers with stride equal to $2$ are utilized to extract information and down-sample the extracted feature map. Meanwhile, the number of channels also doubles successively with the down-sampling. Every convolution layer is followed by batch normalization \cite{Ioffe_2015_arxiv} and rectified linear unit \cite{Krizhevsky_2012_NIPS}. When an image is large, we cannot obtain a $3 \times 1$ vector just by $5$ convolution layers. In order to adapt more images with different sizes, an adaptive average pooling is utilized after $5$ down-sampling convolution layers to down-sample the features to $1 \times 1$ size directly. At last, a fully connected layer synthesizes the whole extracted information into a $3 \times 1$ vector, namely our global atmospheric light $\mathbf{A}$.

\subsection{Network Structure for $\mathcal{T(\cdot)}$}

Before introducing our network structure for $\mathcal{T(\cdot)}$, we first pay attention to the basic unit structures. In \cite{Zhang_2018_CVPR_Shuffle}, Zhang \etal proposed an extremely computation-efficient and performance-preserving ShuffleNet structure. In ShuffleNet units, pointwise group convolution and channel shuffle are used to boost the propagation of information flow. Group convolution helps to encode more information and channel shuffling enables different groups to share cross-group information. The reason why we use ShuffleNet units is that the influence of rain is complex, its sizes and patterns are various. ShuffleNet units make best use of the extracted features to estimate a better transmission.

In order to adapt to the complex rain, revised ShuffleNet units are used in our work, shown in Figure \ref{fig:shuffle_unit}. We deepen the extracted features of single unit by adding another $3 \times 3$ depthwise convolution \cite{Chollet_2016_arxiv}. To reduce the influence of padded $0$ in common convolution, we use symmetrically padded depthwise convolution (SDWConv). Pointwise group convolution and channel shuffling also double to boost information pass.

$\mathcal{T}(\cdot)$ also takes the rainy image $\mathbf{I}$ as input to estimate the transmission $\mathbf{T}$:
\begin{equation}\label{eq:mapping_T}
    \widehat{\mathbf{T}} = \mathcal{T}(\mathbf{\mathbf{I}}),
\end{equation}
where $\widehat{\mathbf{T}}$ is the estimation of $\mathbf{T}$. In the network structure, shown in Figure \ref{fig:whole_net}, we first use a convolution layer to extract feature shallowly. Then two ShuffleNet units which use concatenation method to combine the bottleneck and the shortcut (ShuffleUnit(cat)) are adopted to down-sample the feature map and extract feature. Furthermore, $6$ ShuffleNet units which use addition method to combine the bottleneck and the shortcut (ShuffleUnit(add)) are utilized to deepen our network. Please refer to \cite{Zhang_2018_CVPR_Shuffle} for the details of ShuffleNet unit. At last, we up-sample the feature map to original size and convolution layers are followed to fuse multi-channel features. By a Sigmoid activation function, the transmission can be obtained.

Based on the above definition, the rain-removed result $\widehat{\mathbf{J}}$ can be obtained by:
\begin{equation}\label{eq:mapping_DTA}
    \widehat{\mathbf{J}} = (\mathbf{I} - (1-\mathcal{T}(\mathbf{I})) \circ \mathcal{A}(\mathbf{I})) \oslash \mathcal{T}(\mathbf{I}).
\end{equation}

\subsection{Training Loss}

Our DEMO-Net is trained by two steps. If we utilize $\widetilde{\mathbf{A}}$ to denote the global atmospheric light estimated initially in Section \ref{sec:atmospheric_light}, we first train $\mathcal{A}(\cdot)$ on the given training samples $\{ (\mathbf{I}_{t}, \widetilde{\mathbf{A}}_{t}) \}^{N}_{t=1}$. Then $\mathcal{T}(\cdot)$ is trained on the training samples $\{ (\mathbf{I}_{t}, \mathbf{J}_{t}) \}^{N}_{t=1}$ ($\mathbf{J}_{t}$ is the ground-truth of $\mathbf{I}_{t}$) and $\mathcal{A}(\cdot)$ is fine-tuned during this training process.

\begin{figure}[!t]
\begin{center}
\begin{minipage}{0.32\linewidth}
\centering{\includegraphics[width=1\linewidth]{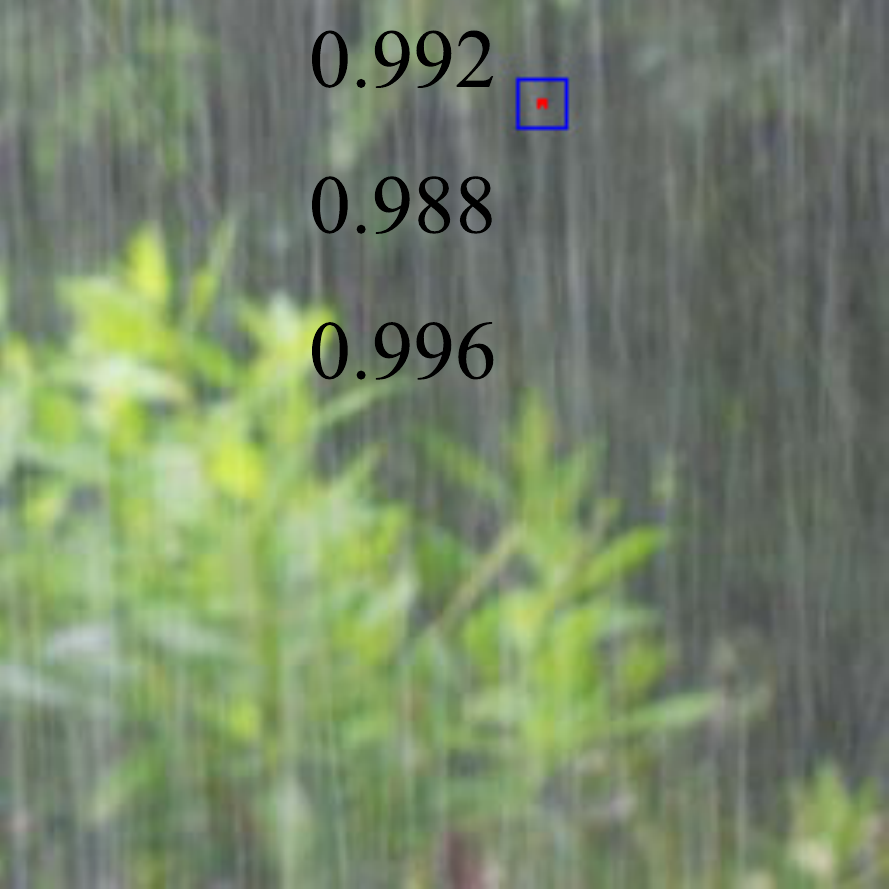}}
\end{minipage}
\hfill
\begin{minipage}{0.32\linewidth}
\centering{\includegraphics[width=1\linewidth]{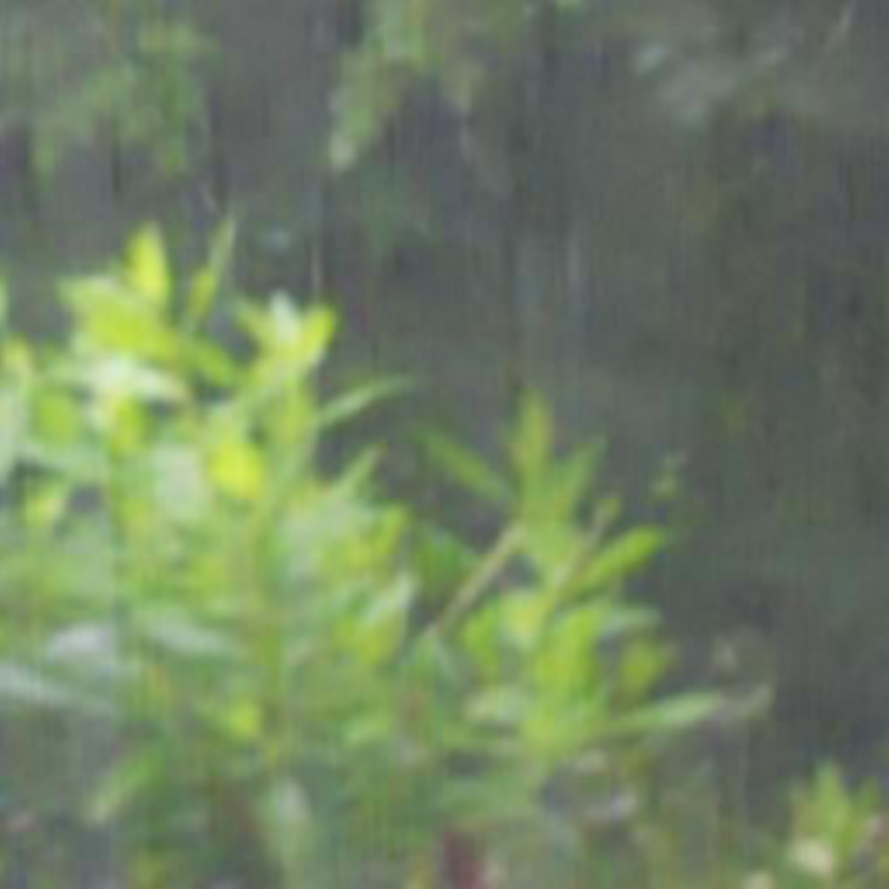}}
\end{minipage}
\hfill
\begin{minipage}{0.32\linewidth}
\centering{\includegraphics[width=1\linewidth]{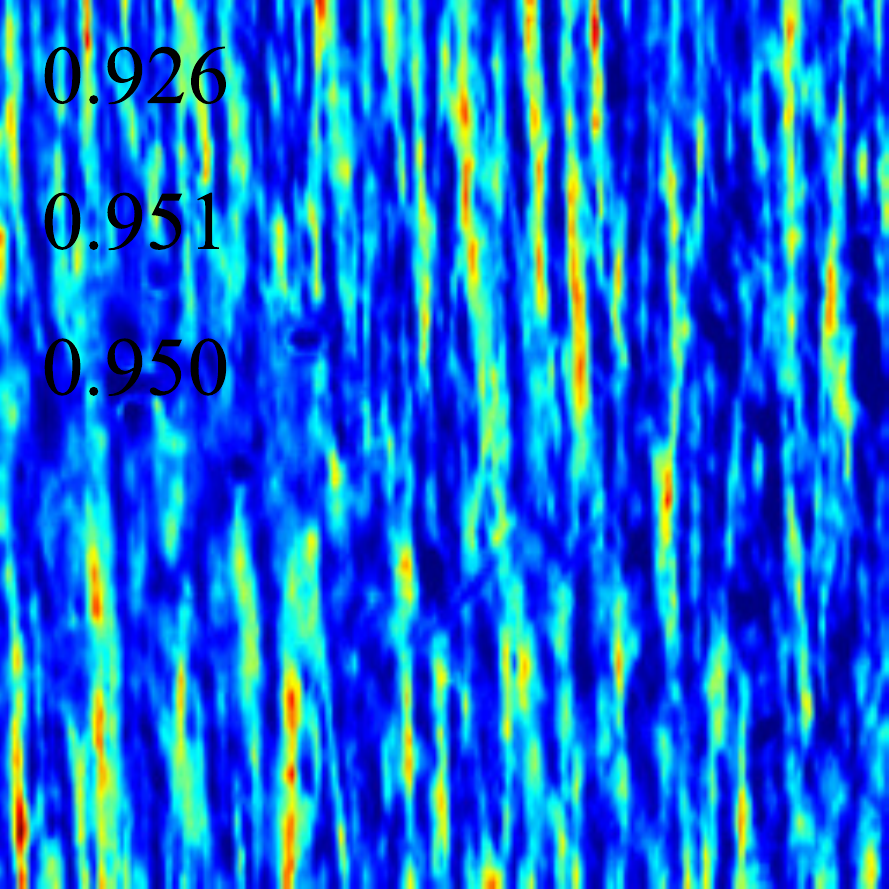}}
\end{minipage}
\vfill
\begin{minipage}{0.32\linewidth}
\centering{\includegraphics[width=1\linewidth]{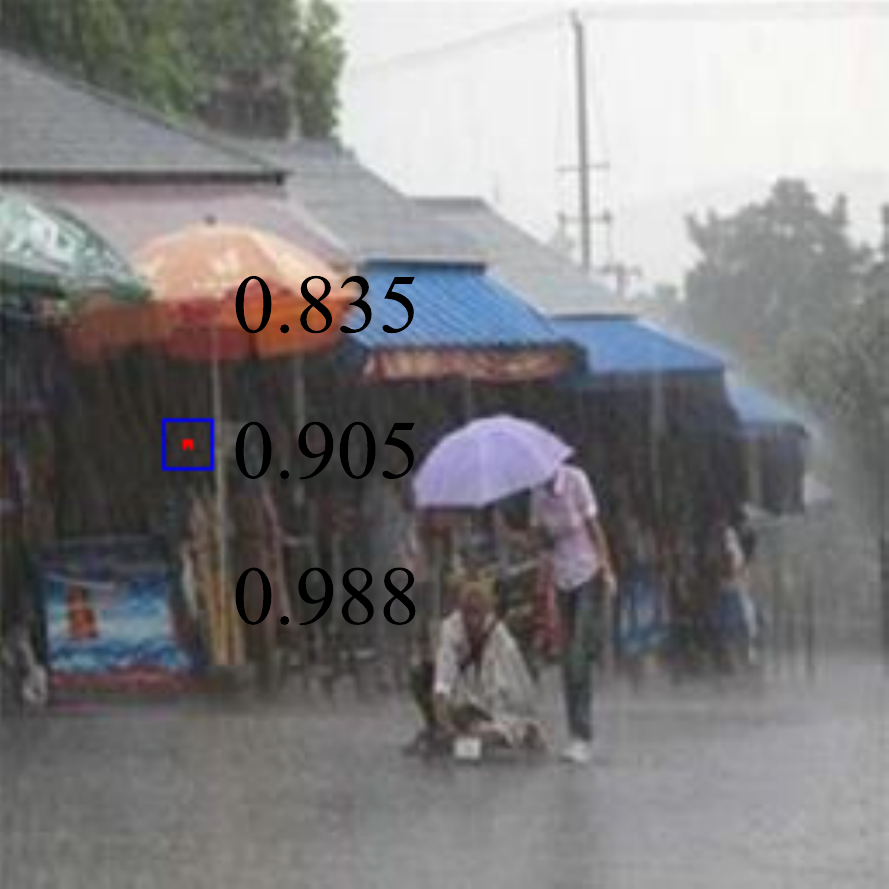}}
\centerline{Rainy images}
\end{minipage}
\hfill
\begin{minipage}{0.32\linewidth}
\centering{\includegraphics[width=1\linewidth]{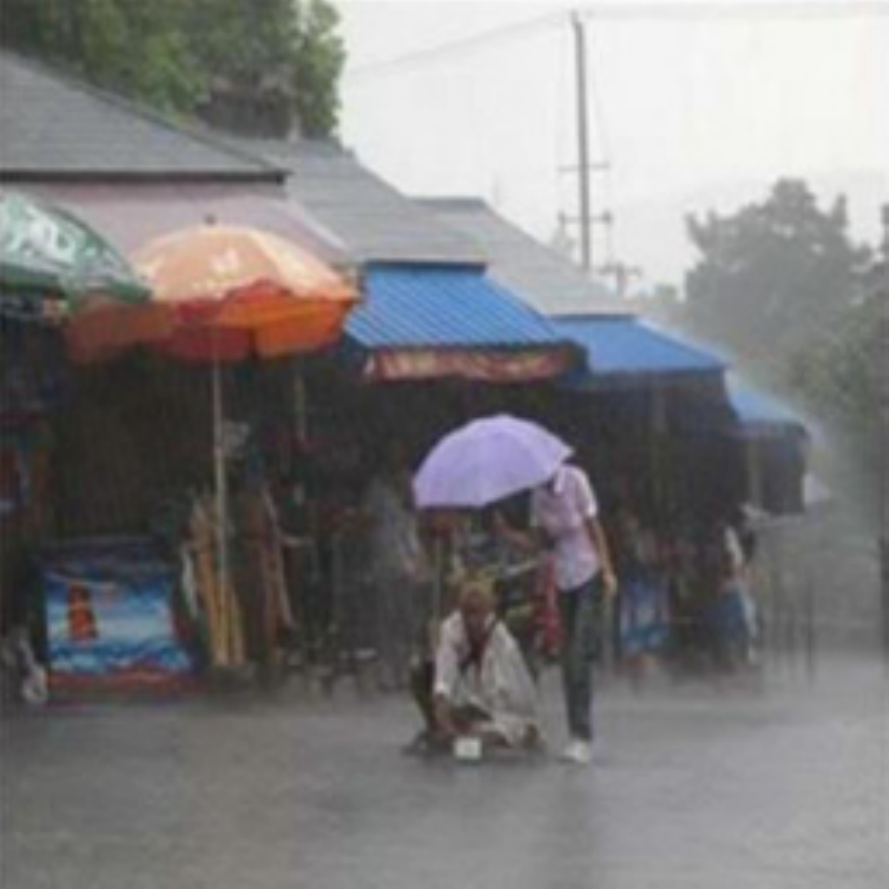}}
\centerline{De-raining results}
\end{minipage}
\hfill
\begin{minipage}{0.32\linewidth}
\centering{\includegraphics[width=1\linewidth]{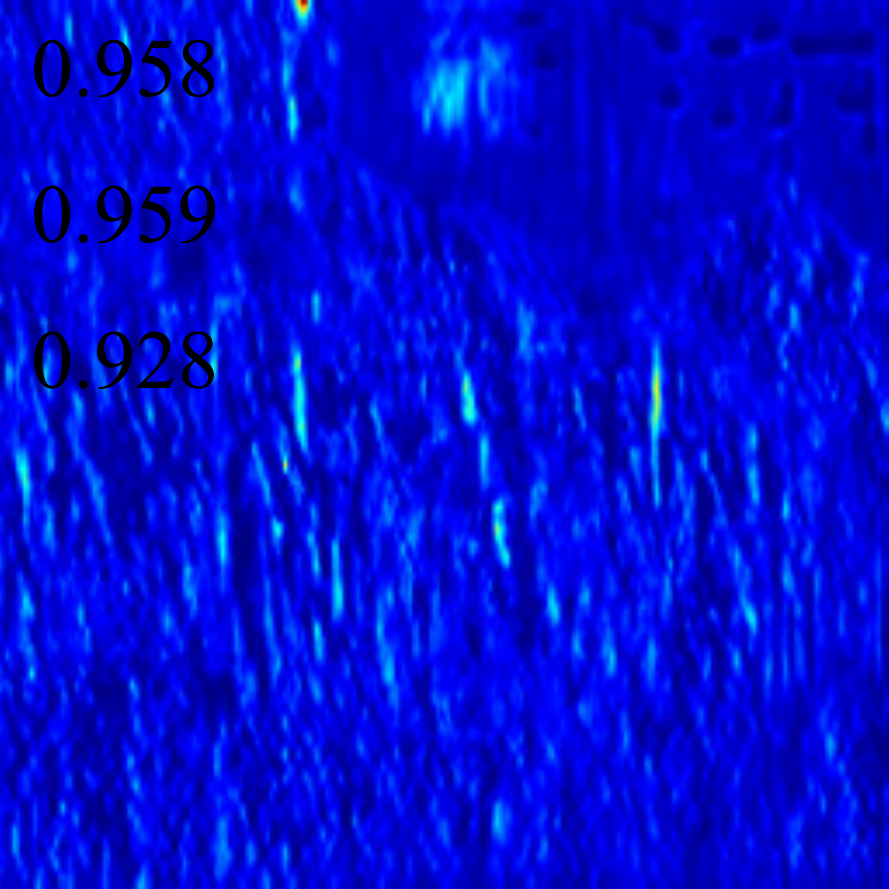}}
\centerline{Transmission}
\end{minipage}
\end{center}
\caption{{In the first column, the red point is the pixel whose value is estimated as $\mathbf{A}$ by the method in Section \ref{sec:atmospheric_light} and we write its value besides the pixel. The vector on images in the last column is the global atmospheric light estimated by $\mathcal{A}(\cdot)$.}}
\label{fig:trans_atm}
\end{figure}

The two-step trainings both utilize MSE loss function:
\begin{equation}\label{eq:loss_A}
    \mathcal{L}_{\mathbf{A}} = \sum^{N}_{t=1} \| \mathcal{A}(\mathbf{I}_{t}) - \widetilde{\mathbf{A}}_{t} \|^{2}_{F},
\end{equation}

\begin{equation}\label{eq:loss_TA}
    \mathcal{L} = \sum^{N}_{t=1} \| \mathcal{D}(\mathbf{I}_{t}) - \mathbf{J}_{t} \|^{2}_{F},
\end{equation}
where $\mathcal{L}_{\mathbf{A}}$ and $\mathcal{L}$ are losses for $\mathcal{A}(\cdot)$ and $\mathcal{T}(\cdot)$ respectively. Some training details are shown in supplement.

\begin{table}[]
\centering
\caption{PSNR and SSIM comparisons of selected state-of-the-art and our methods on our two datasets.}
\begin{tabular}{c|cc|cc}
\hline
 Baseline& \multicolumn{2}{c|}{Rain-I} & \multicolumn{2}{c}{Rain-II} \\ \hline
 Metric&  PSNR         &    SSIM       &  PSNR         &     SSIM      \\ \hline
 \cite{Fu_2017_CVPR}&    29.10       &    0.873       &     30.01      &     0.895      \\
 \cite{Li_2018_ECCV}&     27.51      &   0.897        &      26.68     &    0.830       \\
 \cite{Zhang_2018_CVPR}&    26.66       &     0.885      &    25.33       &     0.867      \\
 \cite{Yang_2017_CVPR}&    27.69       &   0.851        &    29.97       &    0.893       \\ \hline
 Ours &  \textbf{31.35}        &   \textbf{0.898}        &     \textbf{34.41}      &     \textbf{0.936}      \\ \hline
\end{tabular}
\label{tab:psnr_ssim_comparison}
\end{table}

\begin{table*}[]
\centering
\caption{Average complexity comparisons of selected methods and our methods on our two datasets. The image size is $512 \times 512$.}
\begin{tabular}{c|ccccc}
\hline
 Methods & \cite{Fu_2017_CVPR} (CPU) & \cite{Li_2018_ECCV} (GPU) & \cite{Zhang_2018_CVPR} (GPU) & \cite{Yang_2017_CVPR} (GPU) & Ours (GPU) \\ \hline
 Time & $4.06s$ & $0.47s$ & $0.06s$ & $1.39s$ & $0.03s$ \\ \hline
\end{tabular}
\label{tab:time_comparison}
\end{table*}

\begin{figure*}[!t]
\begin{center}
\begin{minipage}{0.135\linewidth}
\centering{\includegraphics[width=1\linewidth]{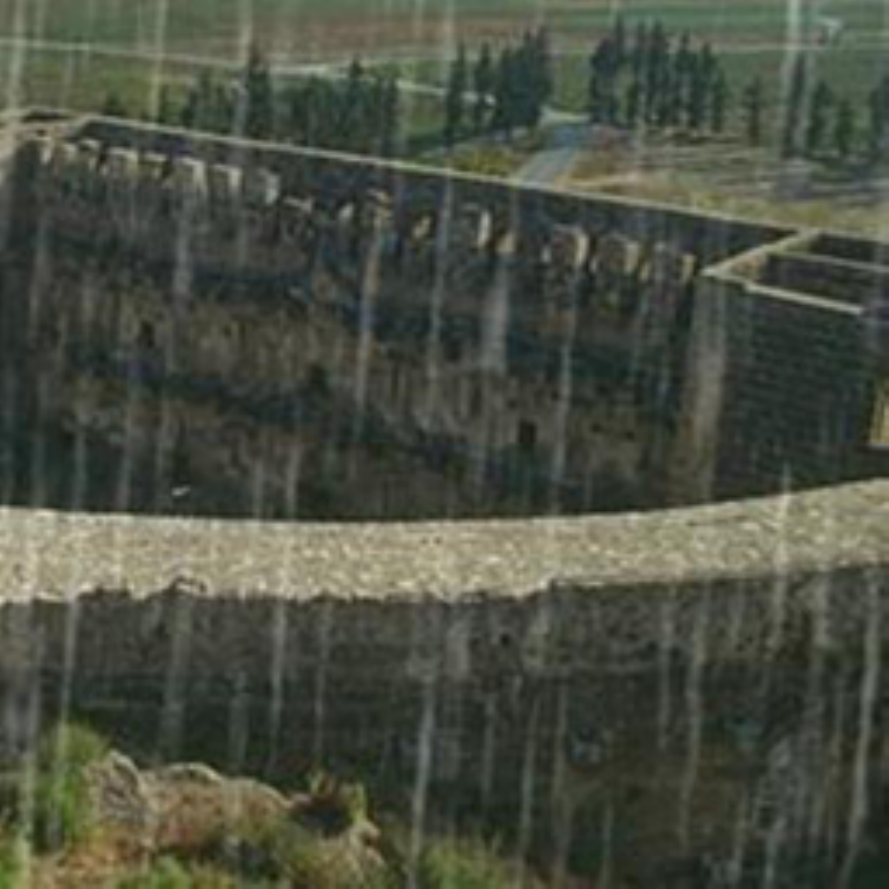}}
\end{minipage}
\hfill
\begin{minipage}{0.135\linewidth}
\centering{\includegraphics[width=1\linewidth]{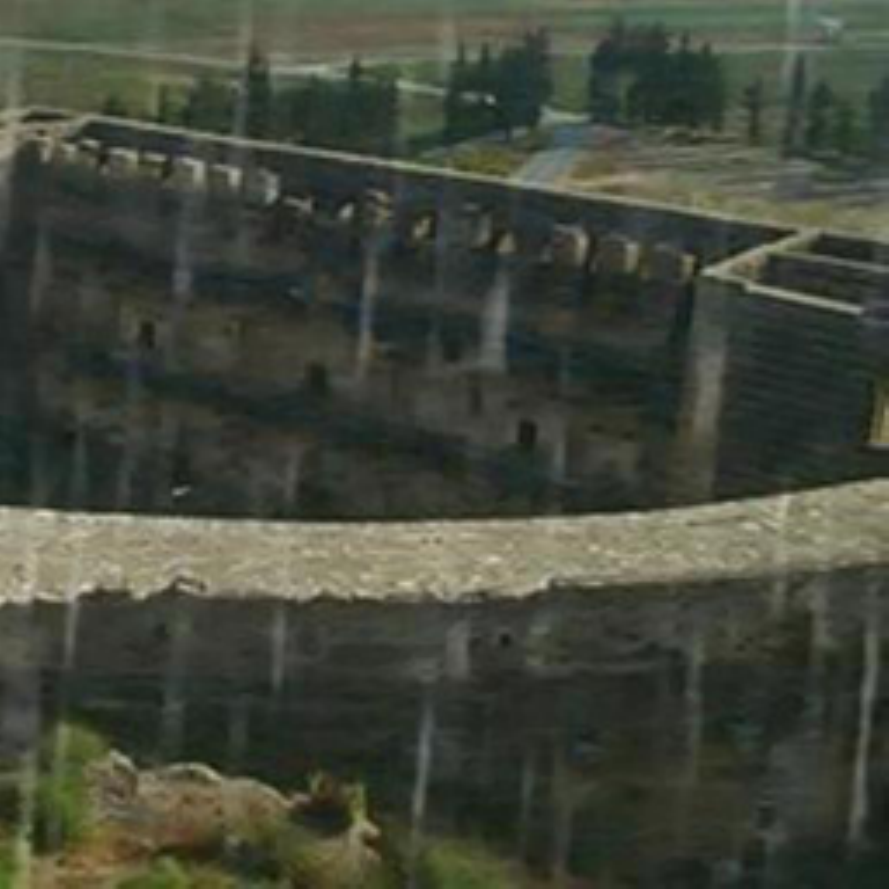}}
\end{minipage}
\hfill
\begin{minipage}{0.135\linewidth}
\centering{\includegraphics[width=1\linewidth]{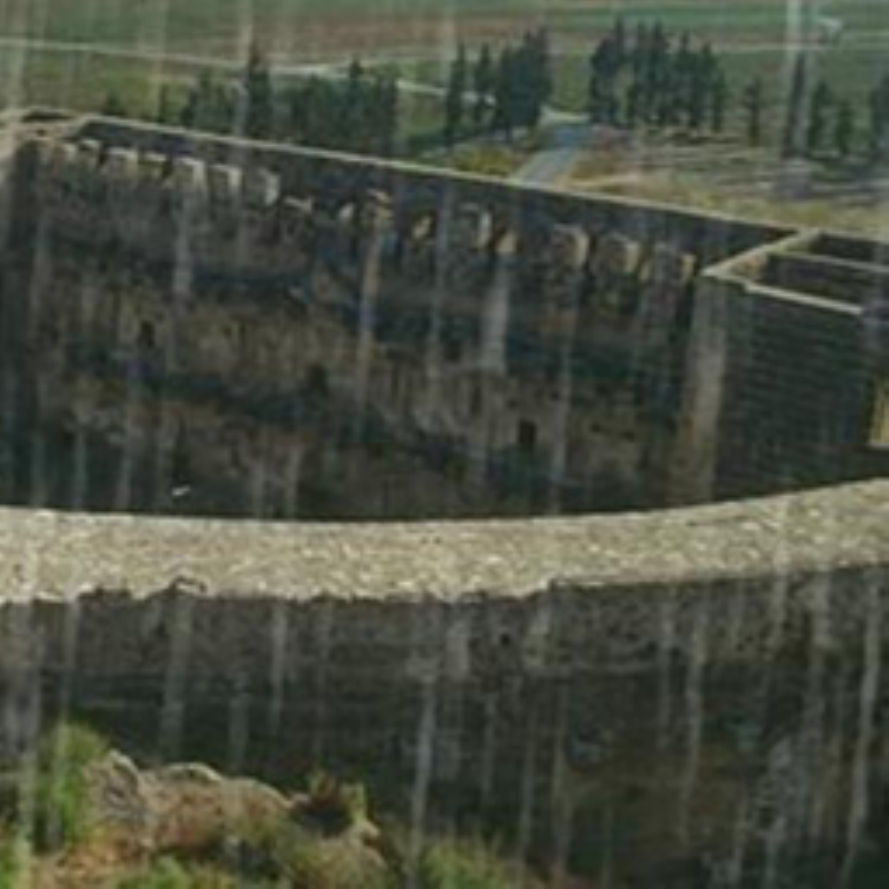}}
\end{minipage}
\hfill
\begin{minipage}{0.135\linewidth}
\centering{\includegraphics[width=1\linewidth]{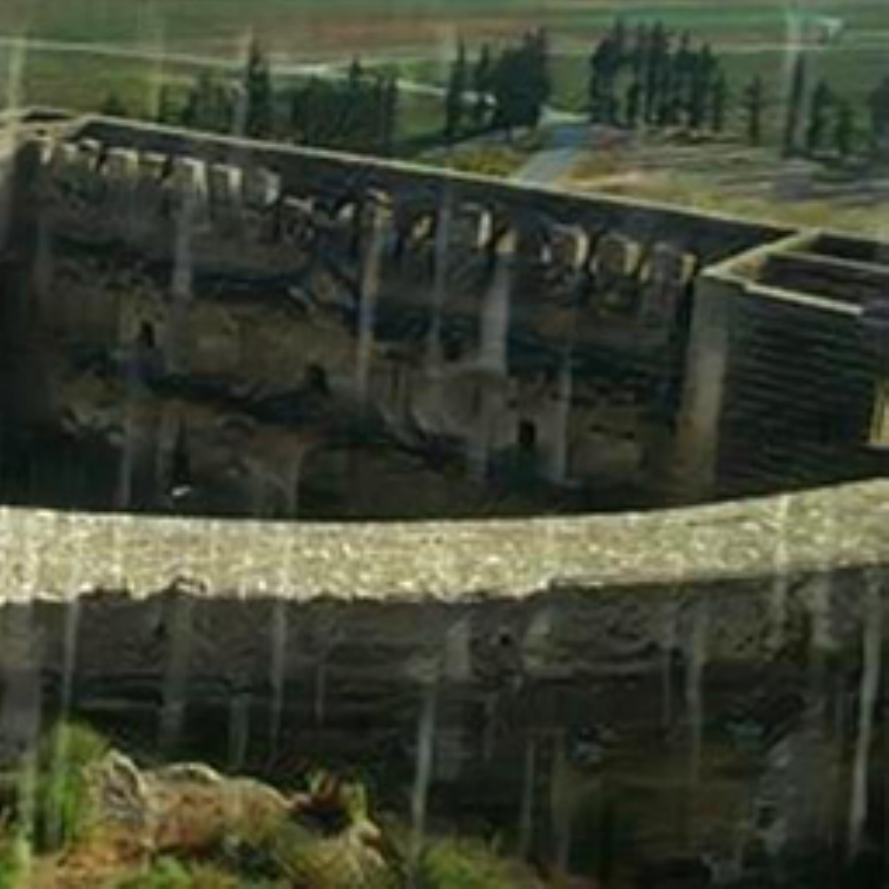}}
\end{minipage}
\hfill
\begin{minipage}{0.135\linewidth}
\centering{\includegraphics[width=1\linewidth]{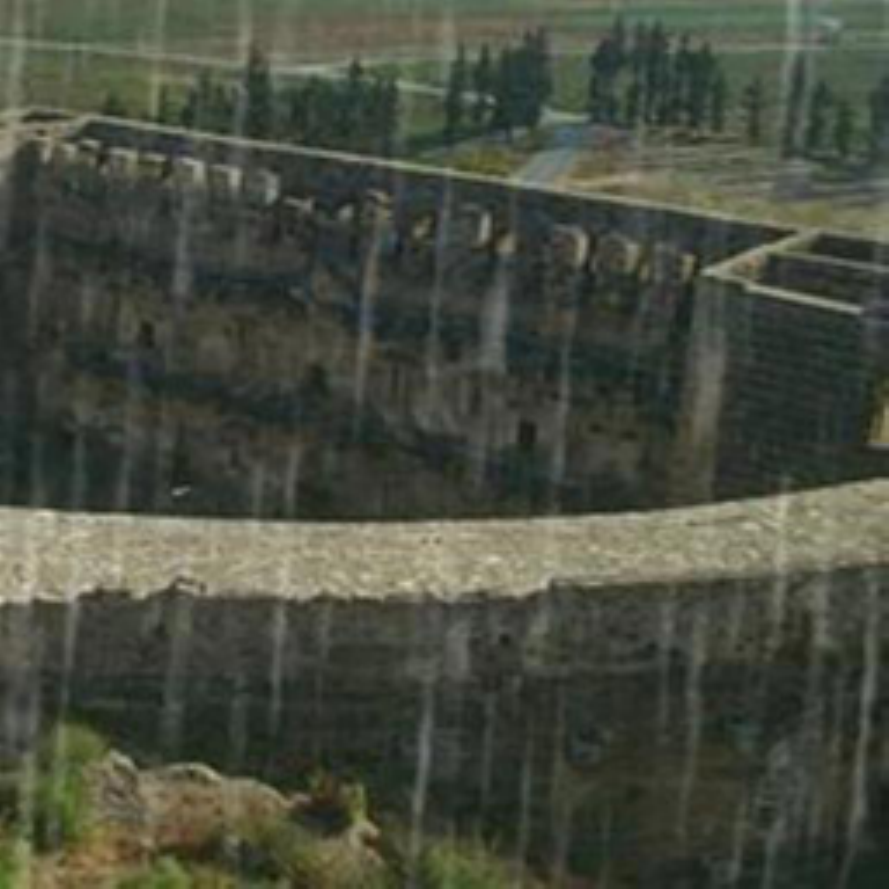}}
\end{minipage}
\hfill
\begin{minipage}{0.135\linewidth}
\centering{\includegraphics[width=1\linewidth]{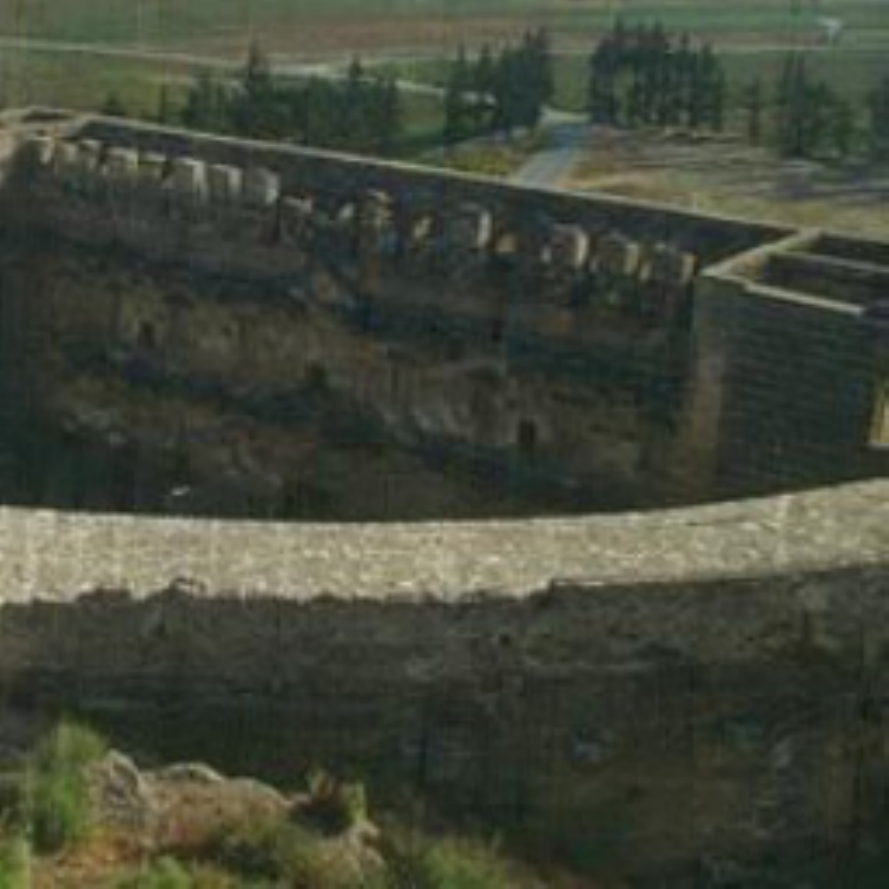}}
\end{minipage}
\hfill
\begin{minipage}{0.135\linewidth}
\centering{\includegraphics[width=1\linewidth]{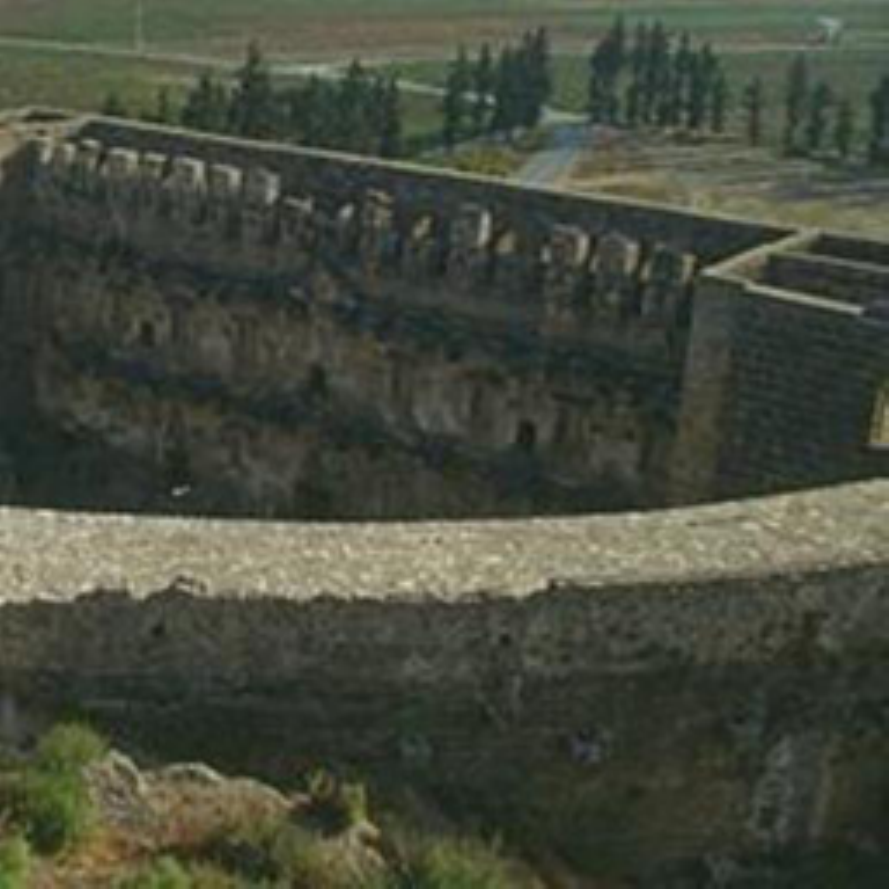}}
\end{minipage}
\vfill
\begin{minipage}{0.135\linewidth}
\centering{\includegraphics[width=1\linewidth]{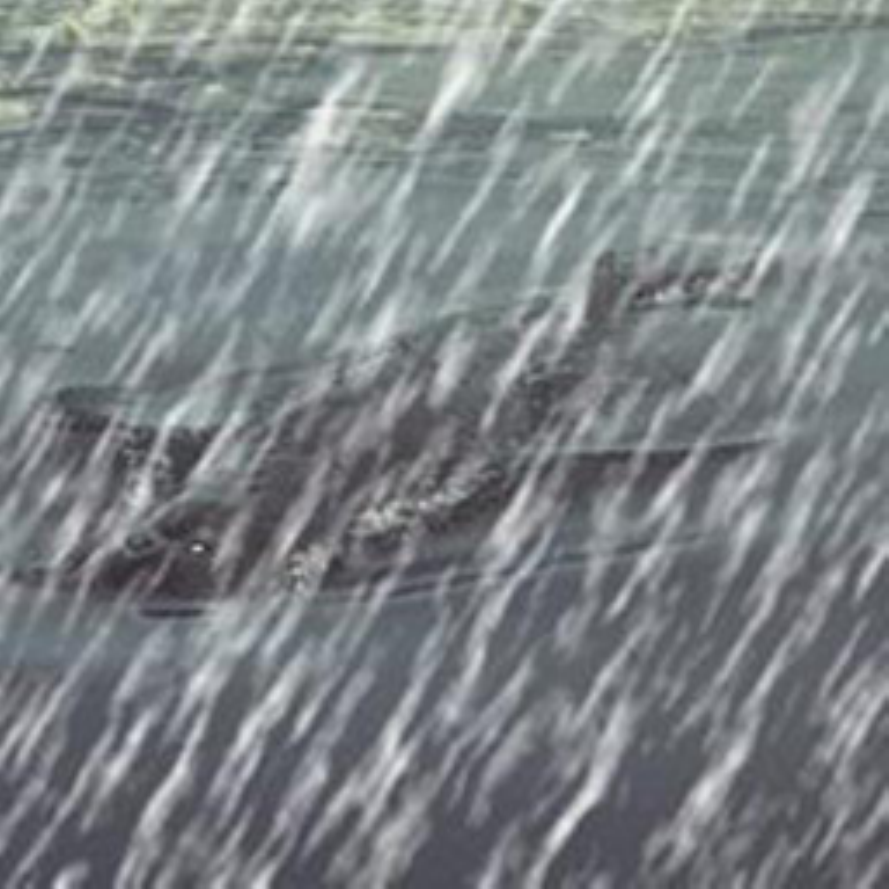}}
\centerline{Input}
\end{minipage}
\hfill
\begin{minipage}{0.135\linewidth}
\centering{\includegraphics[width=1\linewidth]{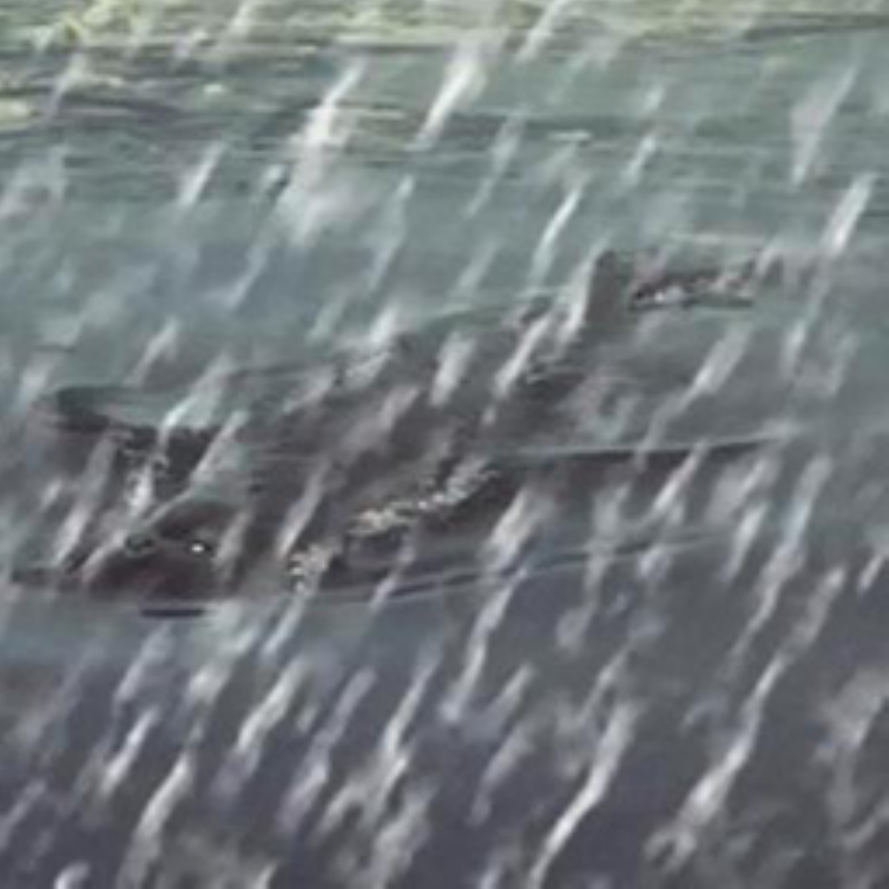}}
\centerline{\cite{Fu_2017_CVPR}}
\end{minipage}
\hfill
\begin{minipage}{0.135\linewidth}
\centering{\includegraphics[width=1\linewidth]{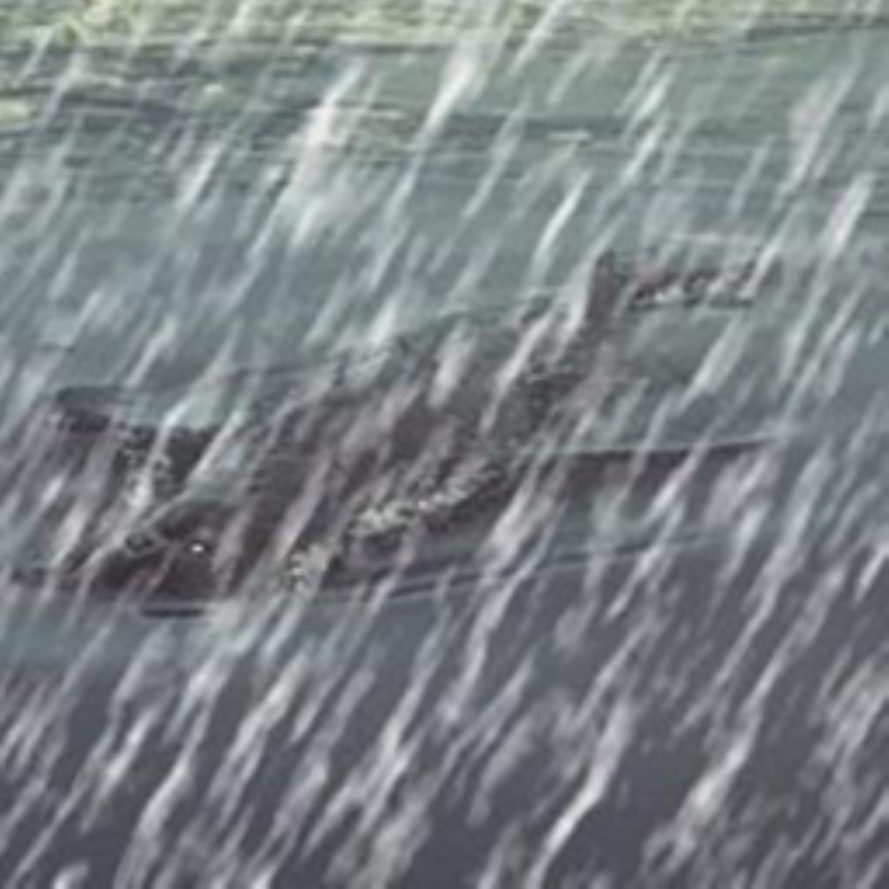}}
\centerline{\cite{Li_2018_ECCV}}
\end{minipage}
\hfill
\begin{minipage}{0.135\linewidth}
\centering{\includegraphics[width=1\linewidth]{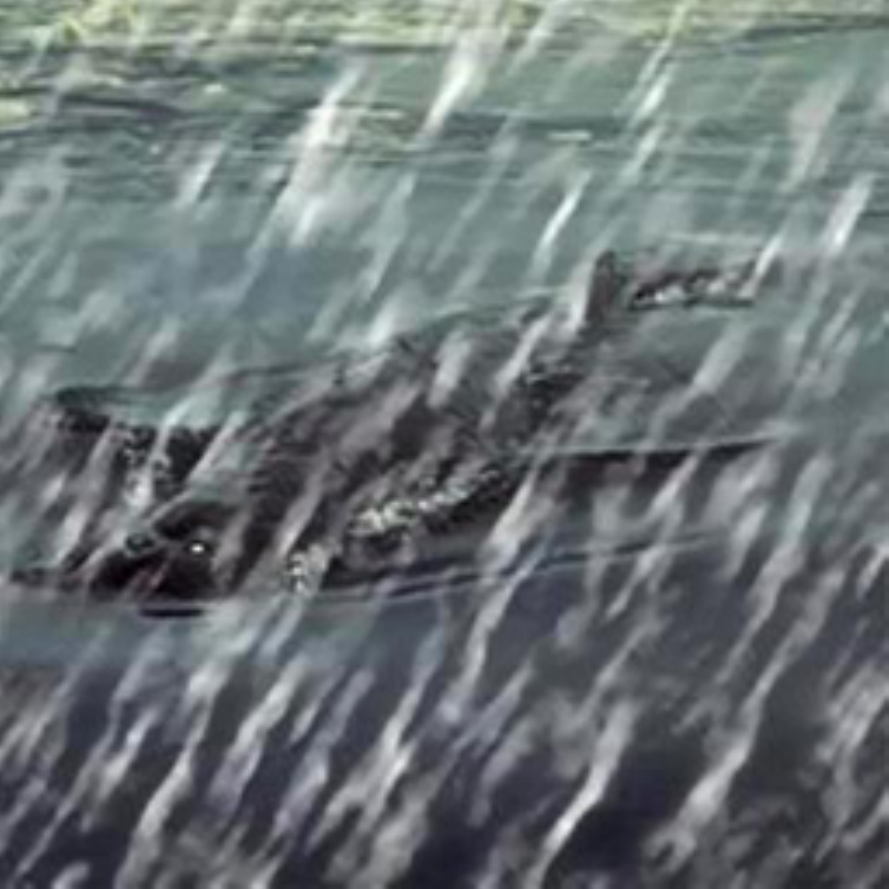}}
\centerline{\cite{Zhang_2018_CVPR}}
\end{minipage}
\hfill
\begin{minipage}{0.135\linewidth}
\centering{\includegraphics[width=1\linewidth]{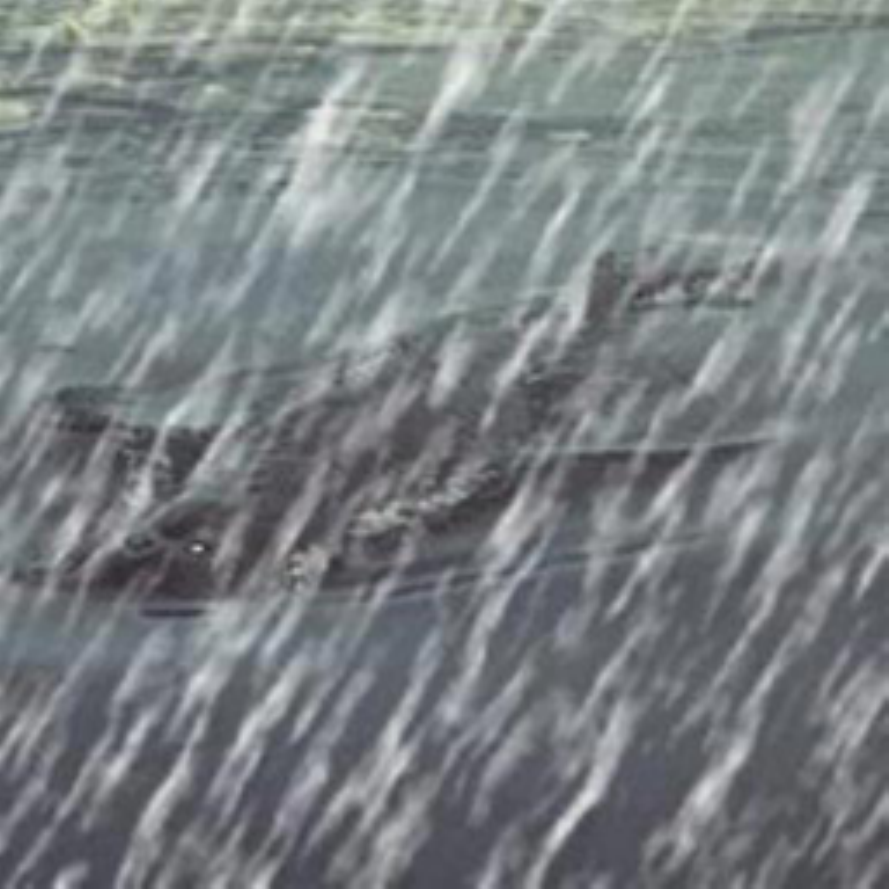}}
\centerline{\cite{Yang_2017_CVPR}}
\end{minipage}
\hfill
\begin{minipage}{0.135\linewidth}
\centering{\includegraphics[width=1\linewidth]{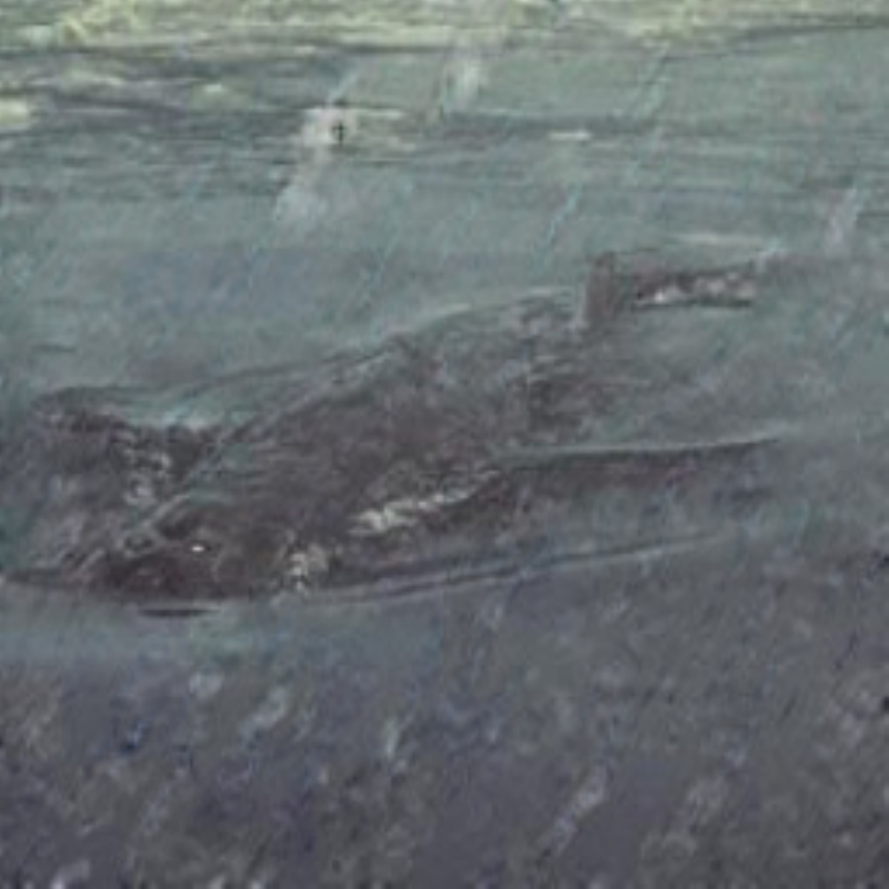}}
\centerline{Ours}
\end{minipage}
\hfill
\begin{minipage}{0.135\linewidth}
\centering{\includegraphics[width=1\linewidth]{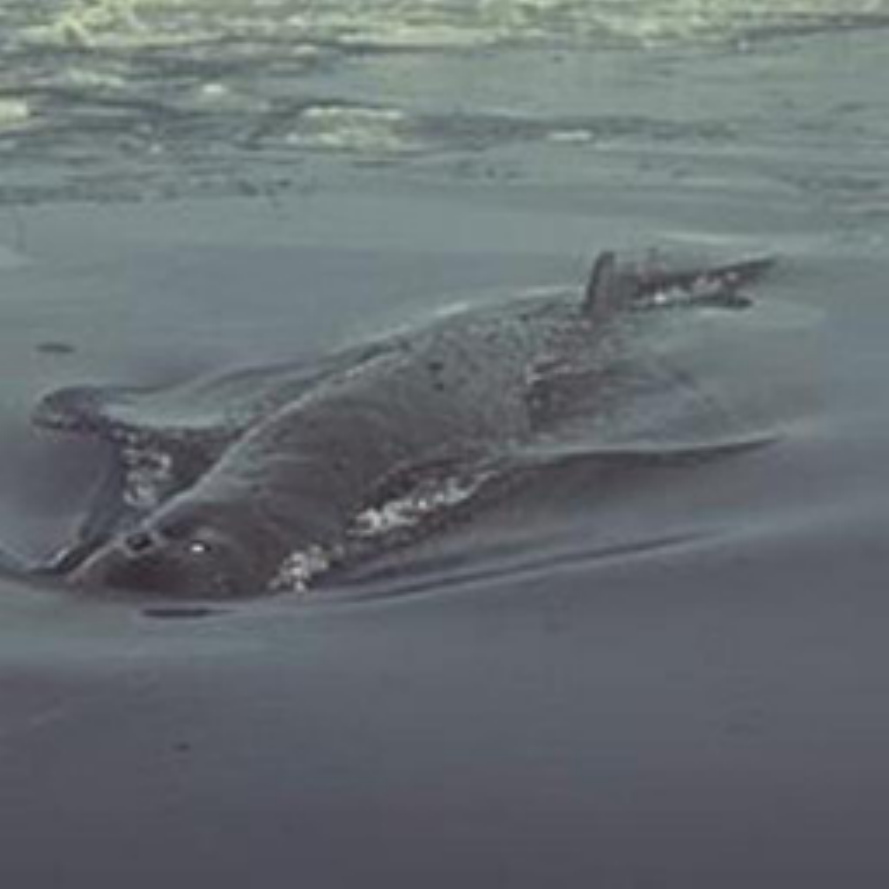}}
\centerline{Ground Truth}
\end{minipage}
\end{center}
\caption{This figure shows the rain-removed results on synthetic rainy images. Here, we show two rainy images which have wide rain streaks whose edges are relatively blurry. The results of other synthetic rainy images are in supplement.}
\label{fig:synthetic_compare}
\end{figure*}

\begin{figure*}[!t]
\begin{center}
\begin{minipage}{0.16\linewidth}
\centering{\includegraphics[width=1\linewidth]{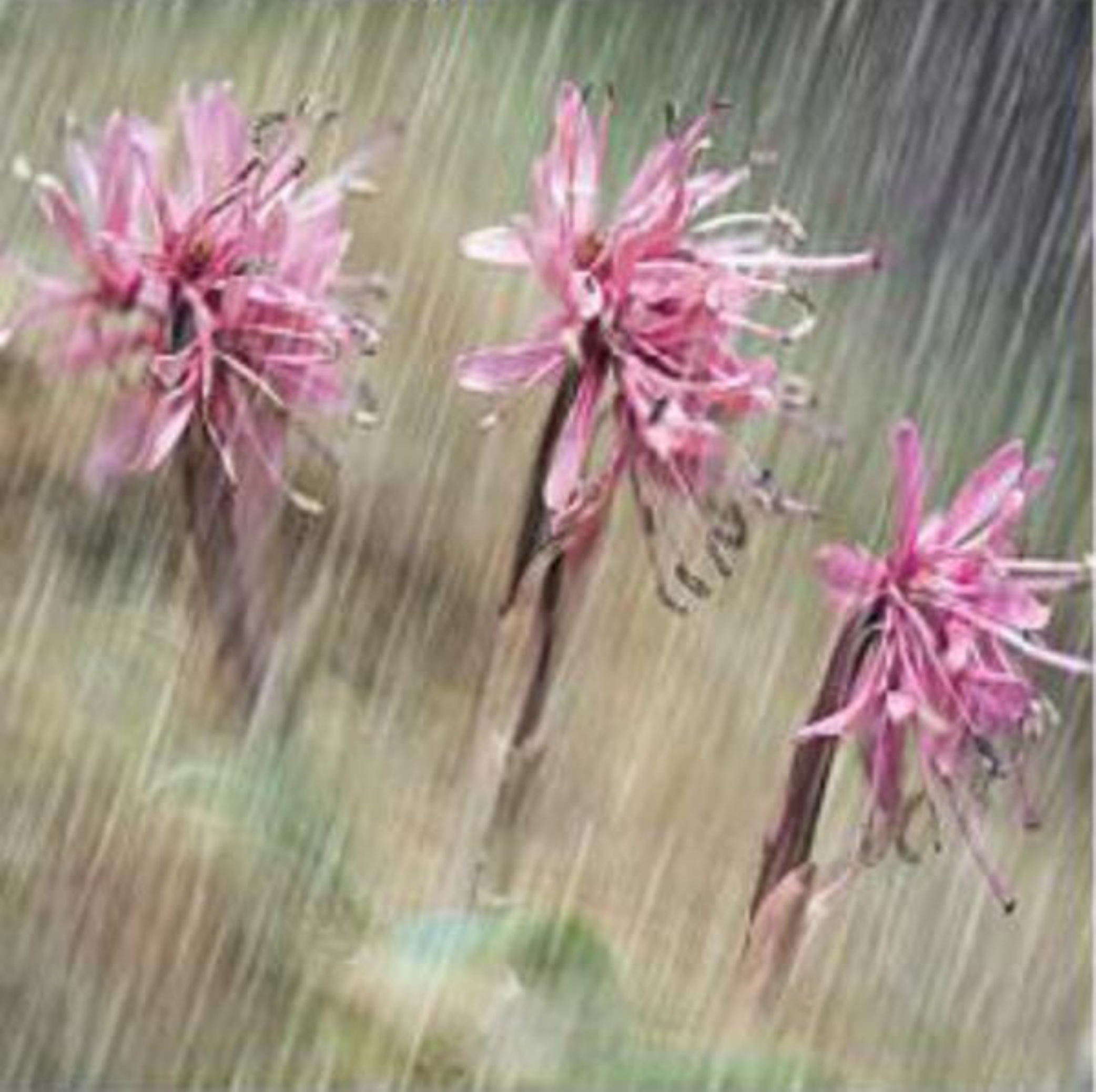}}
\end{minipage}
\hfill
\begin{minipage}{0.16\linewidth}
\centering{\includegraphics[width=1\linewidth]{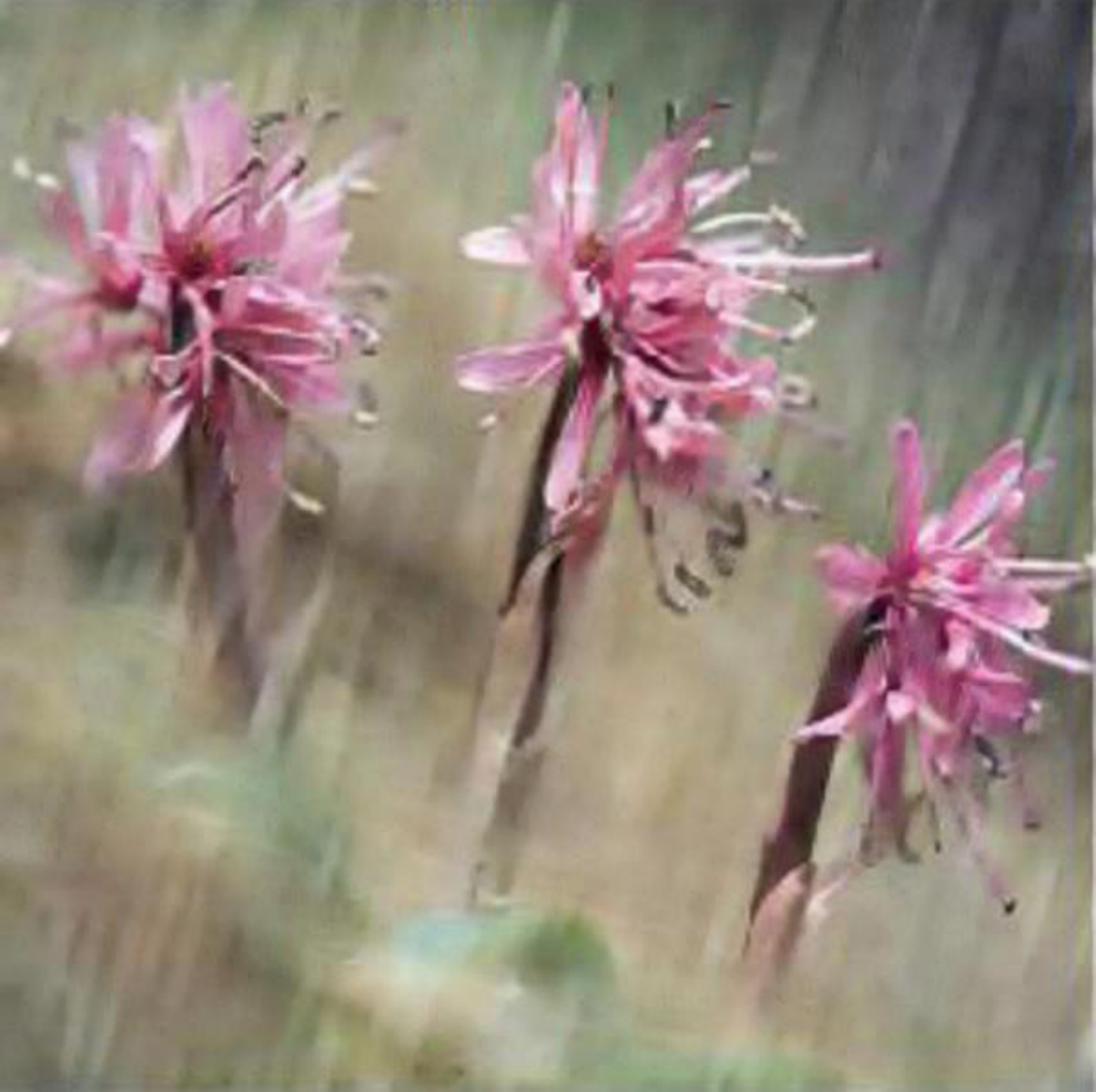}}
\end{minipage}
\hfill
\begin{minipage}{0.16\linewidth}
\centering{\includegraphics[width=1\linewidth]{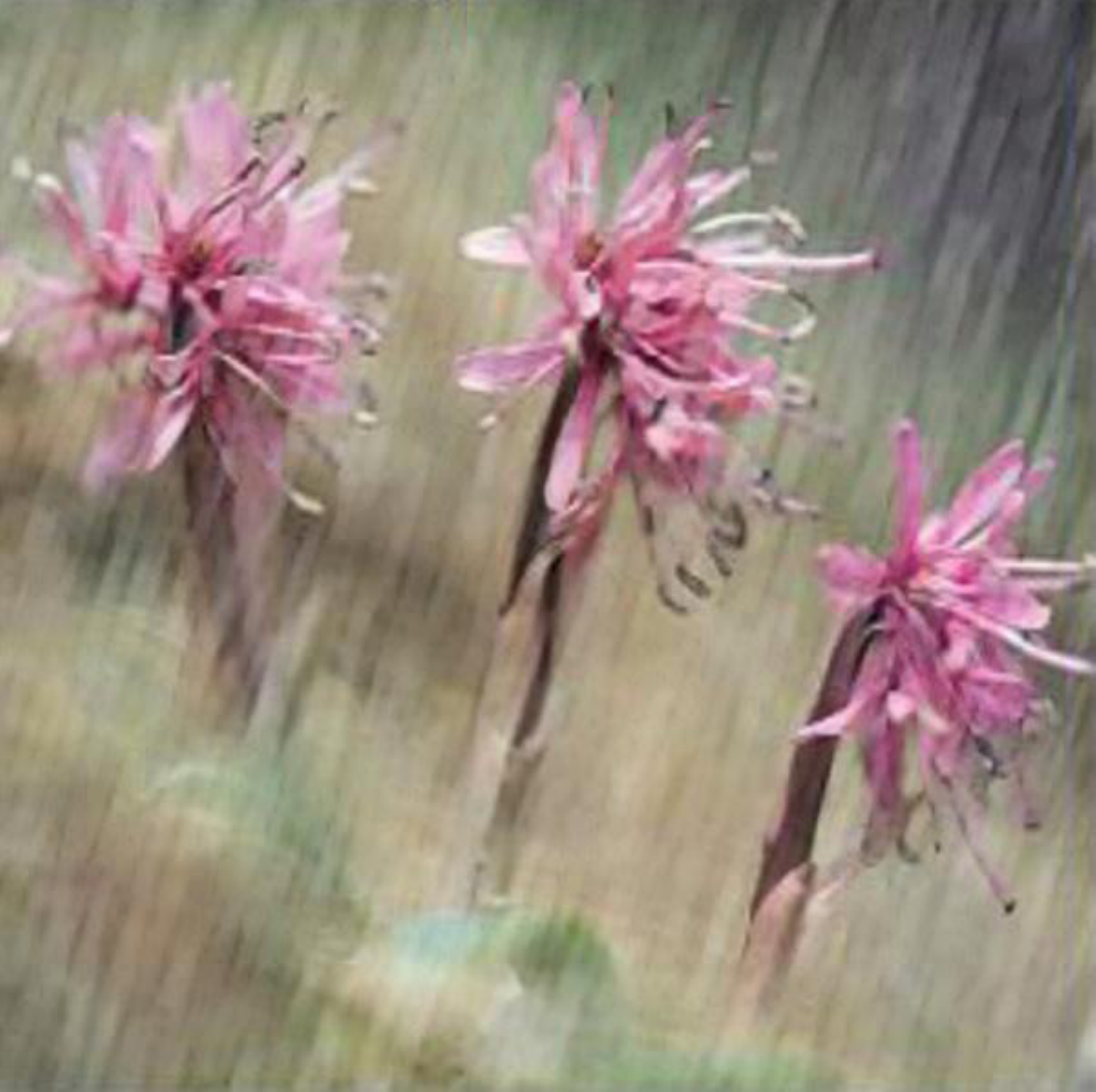}}
\end{minipage}
\hfill
\begin{minipage}{0.16\linewidth}
\centering{\includegraphics[width=1\linewidth]{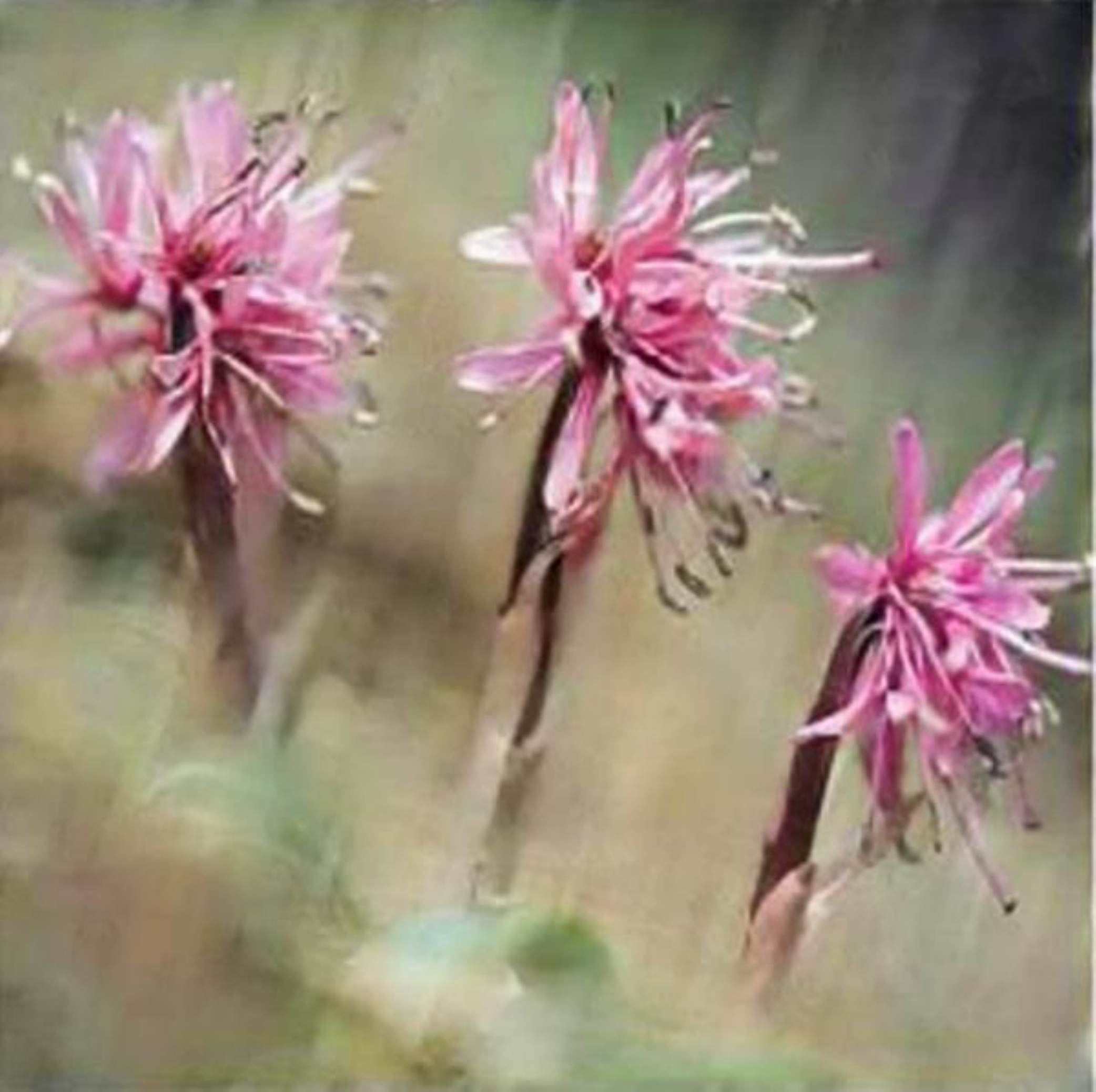}}
\end{minipage}
\hfill
\begin{minipage}{0.16\linewidth}
\centering{\includegraphics[width=1\linewidth]{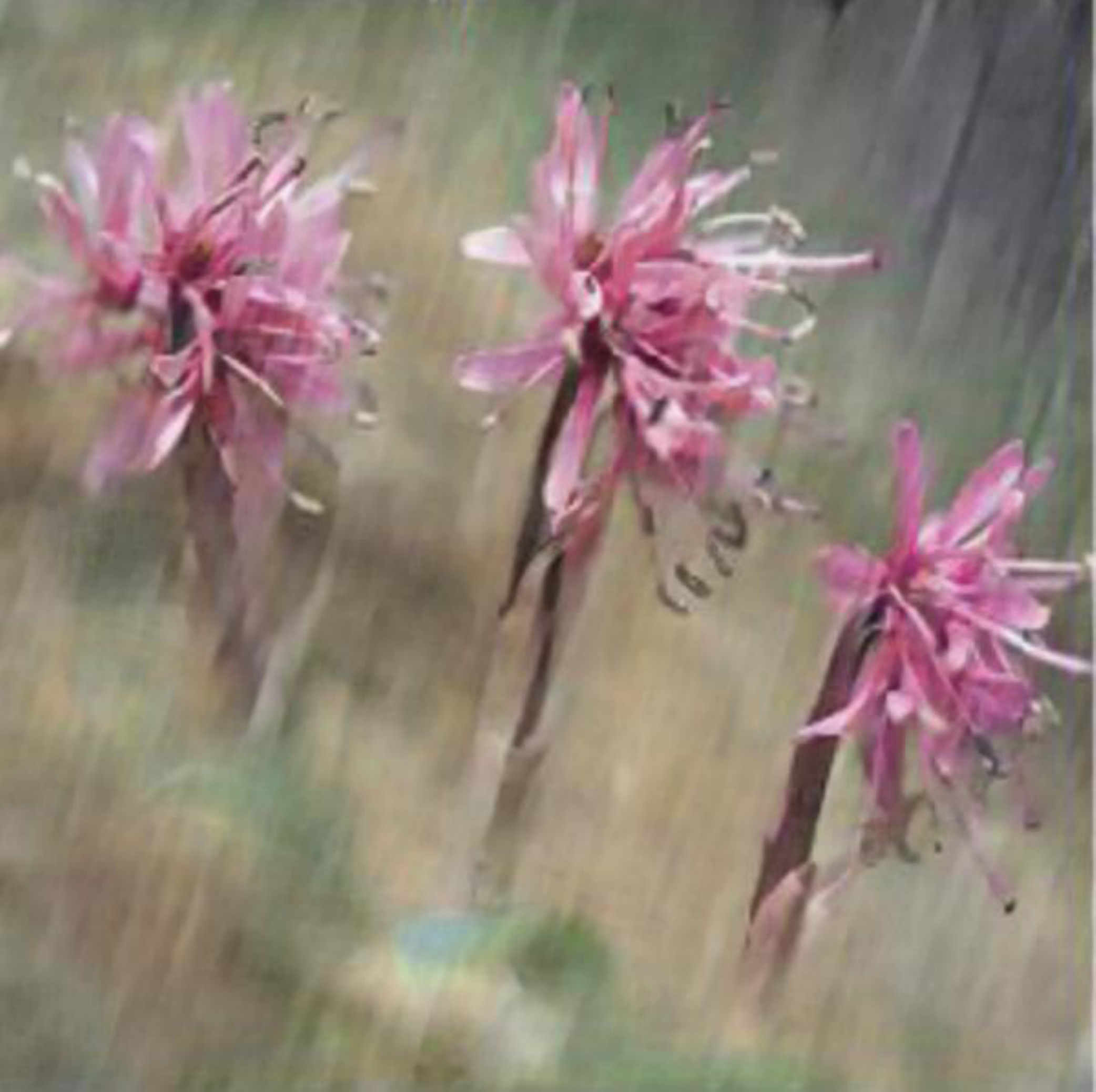}}
\end{minipage}
\hfill
\begin{minipage}{0.16\linewidth}
\centering{\includegraphics[width=1\linewidth]{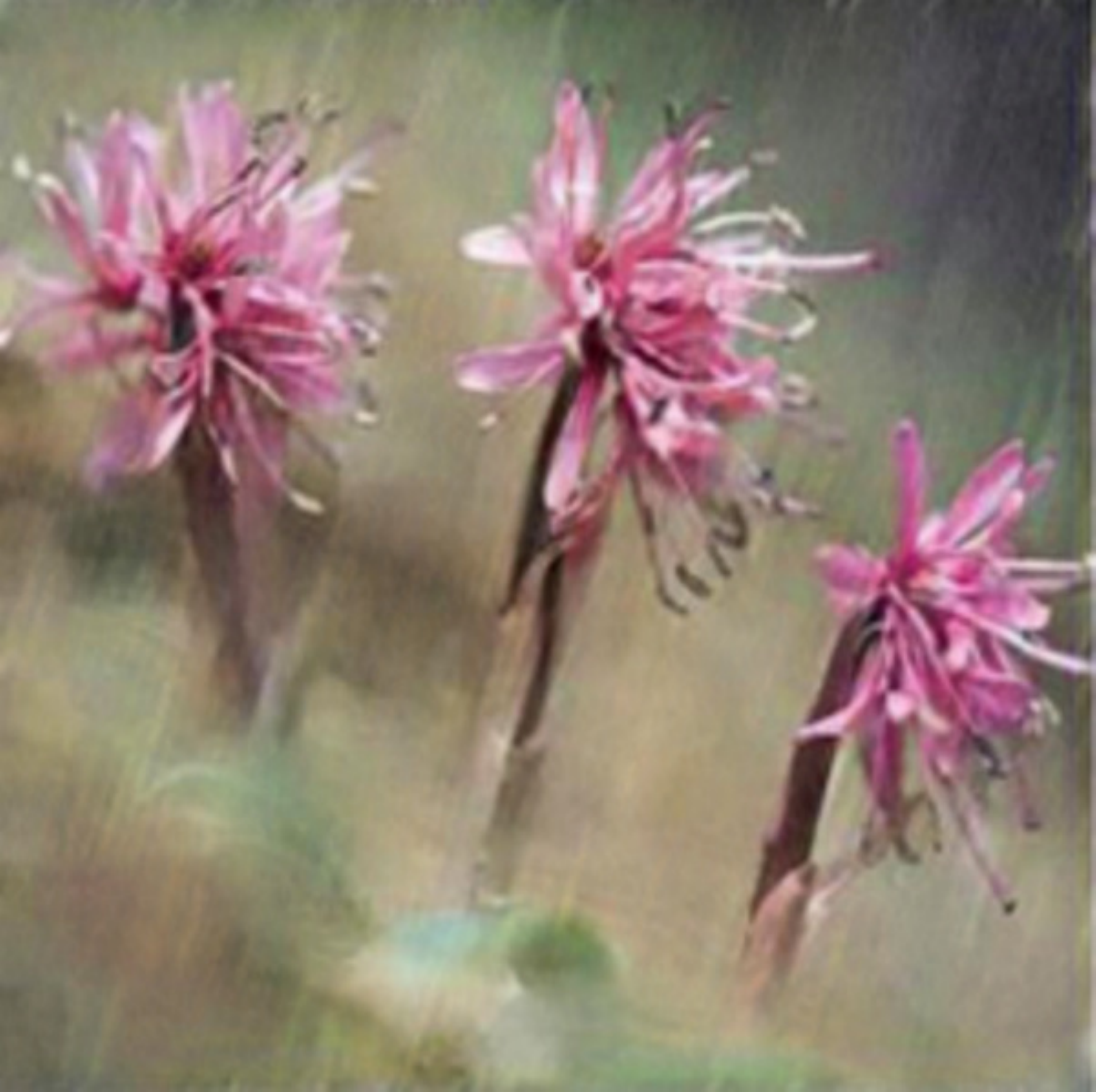}}
\end{minipage}
\vfill
\begin{minipage}{0.16\linewidth}
\centering{\includegraphics[width=1\linewidth]{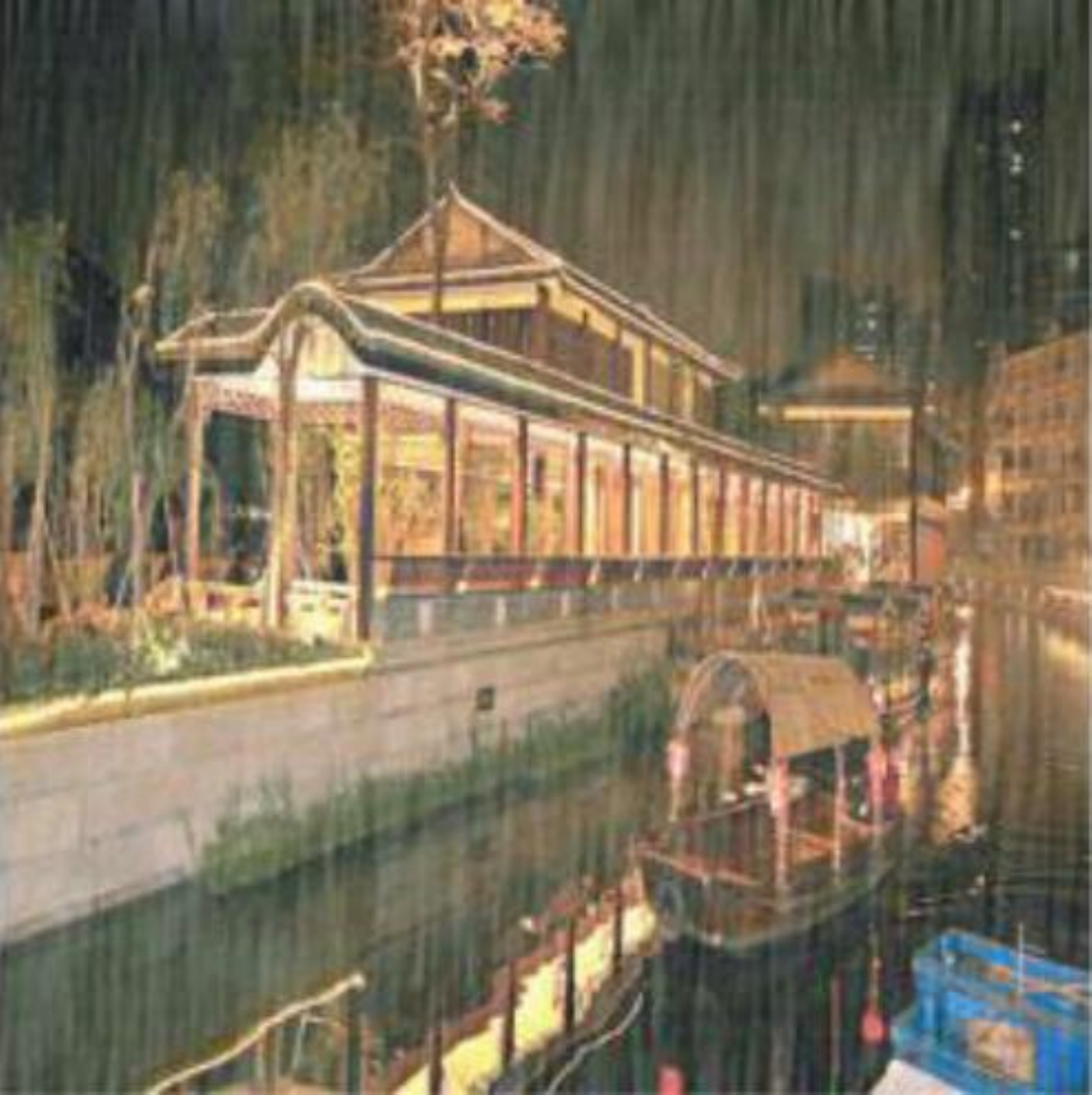}}
\end{minipage}
\hfill
\begin{minipage}{0.16\linewidth}
\centering{\includegraphics[width=1\linewidth]{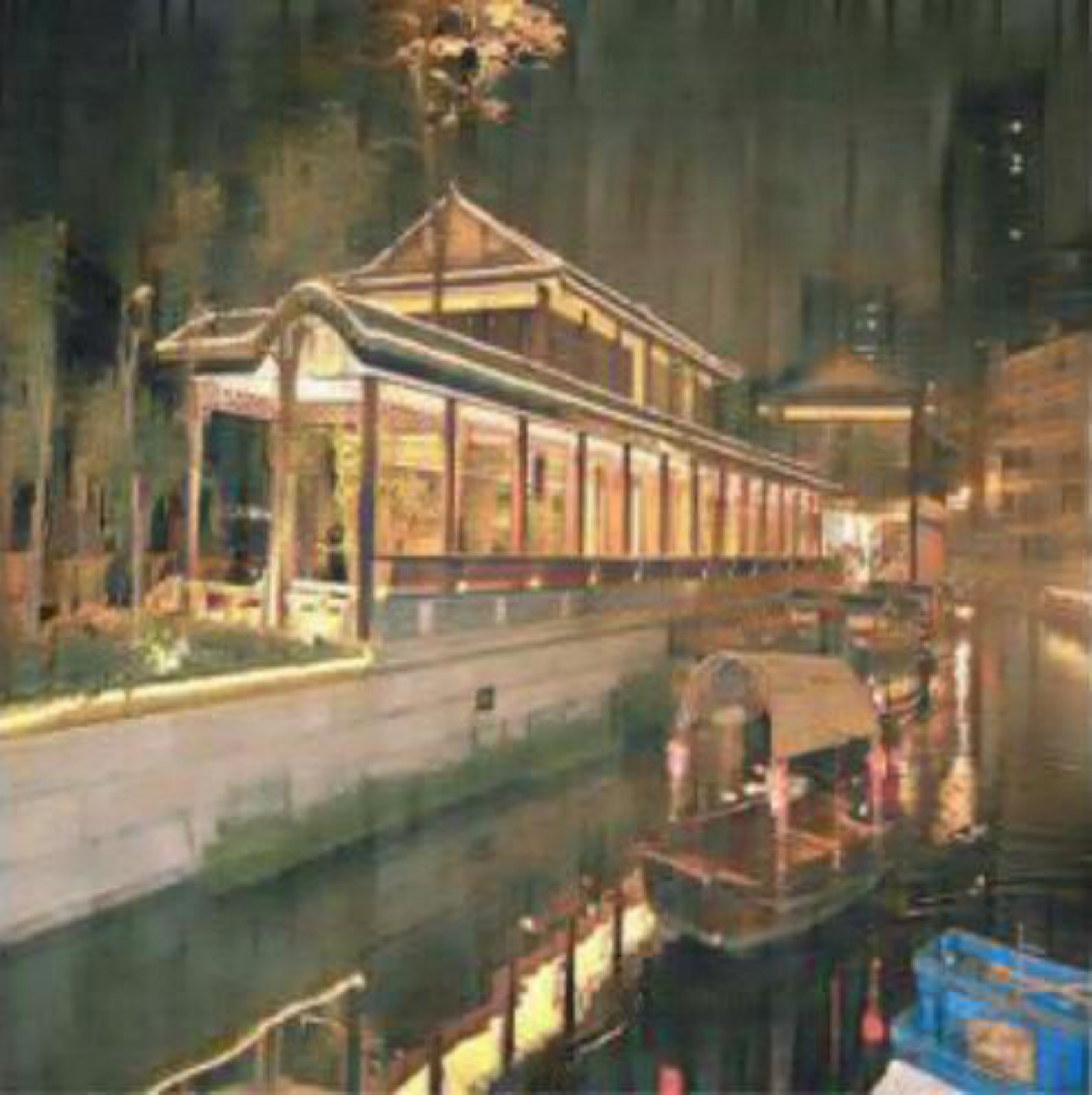}}
\end{minipage}
\hfill
\begin{minipage}{0.16\linewidth}
\centering{\includegraphics[width=1\linewidth]{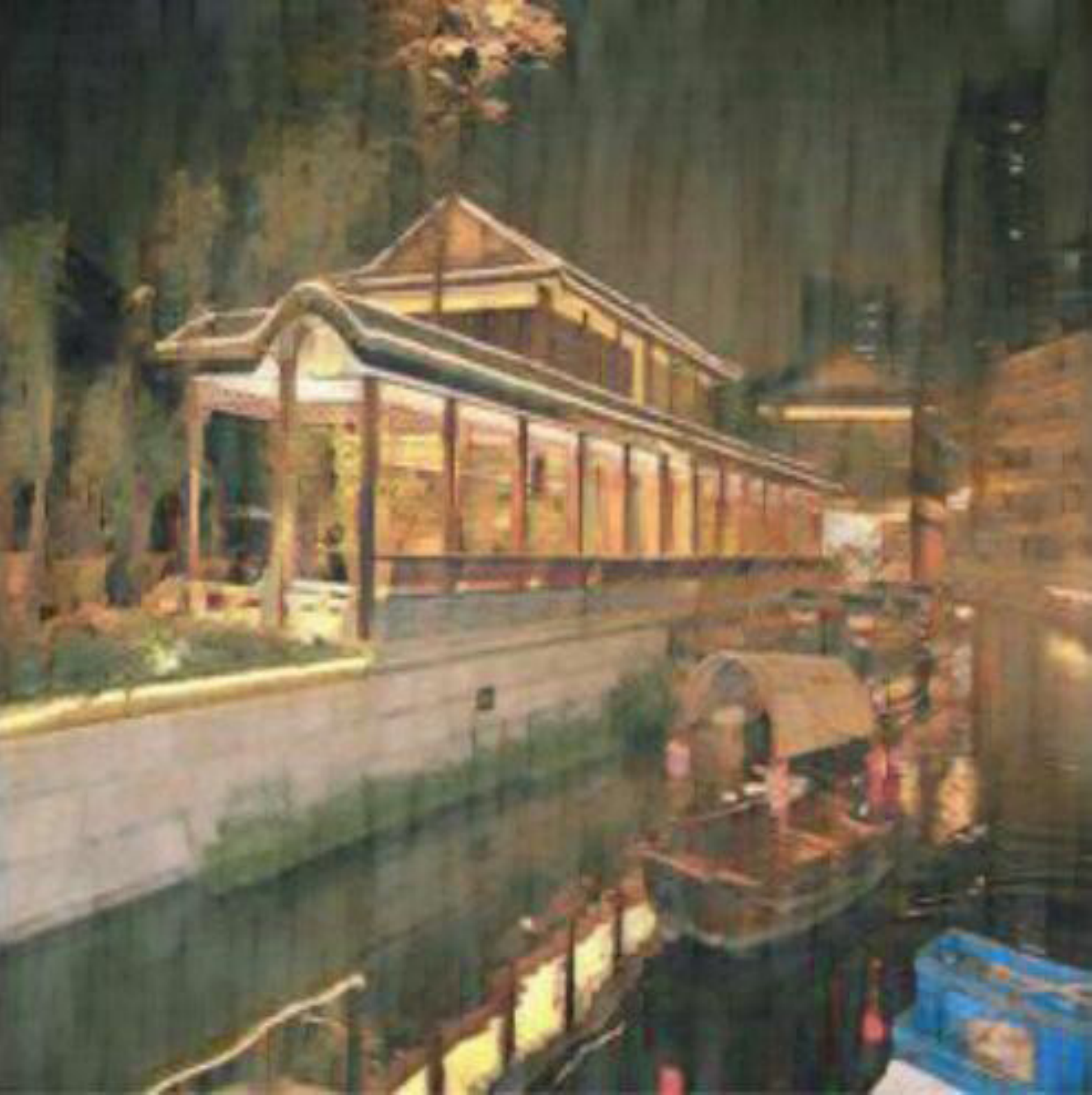}}
\end{minipage}
\hfill
\begin{minipage}{0.16\linewidth}
\centering{\includegraphics[width=1\linewidth]{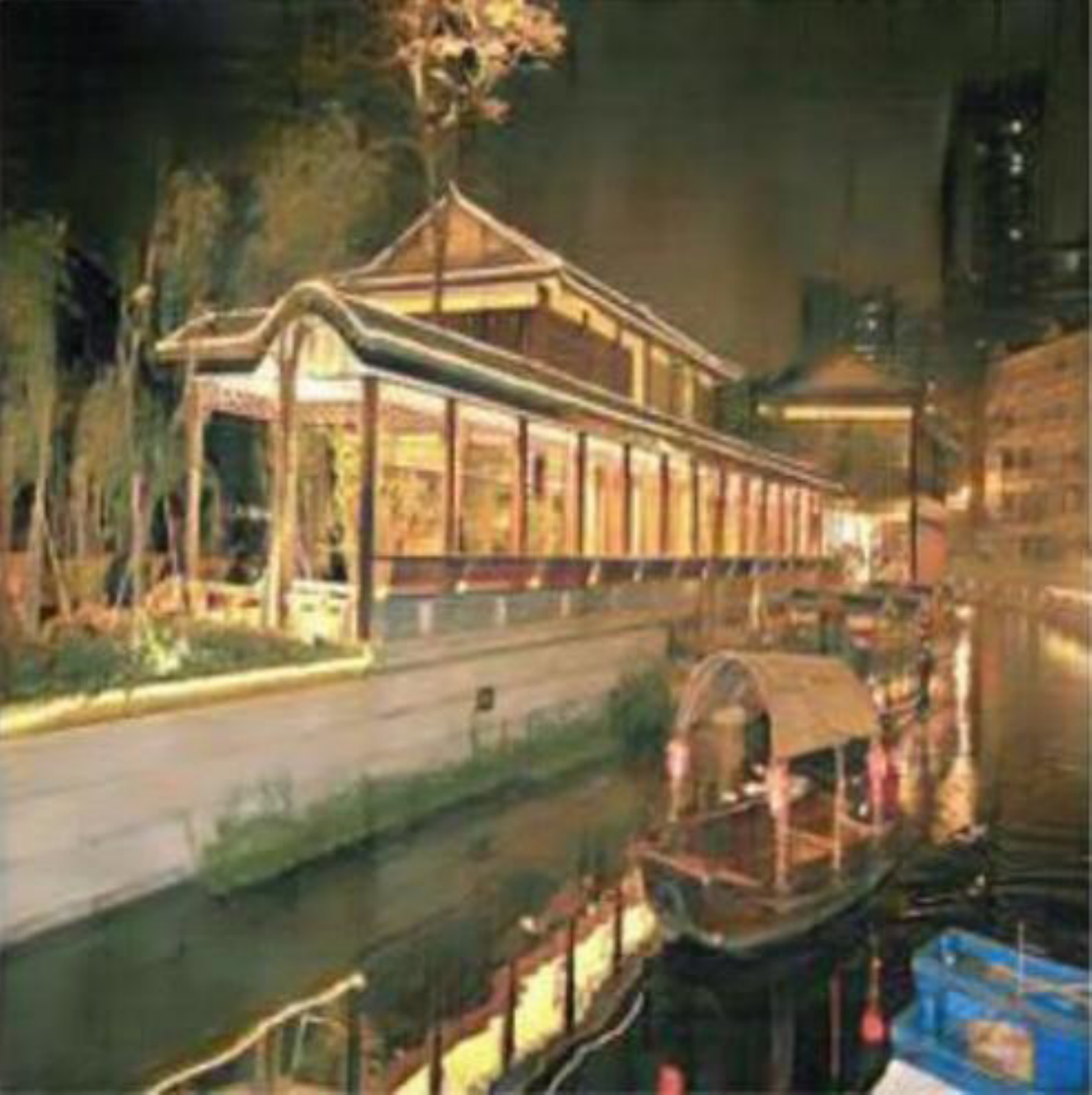}}
\end{minipage}
\hfill
\begin{minipage}{0.16\linewidth}
\centering{\includegraphics[width=1\linewidth]{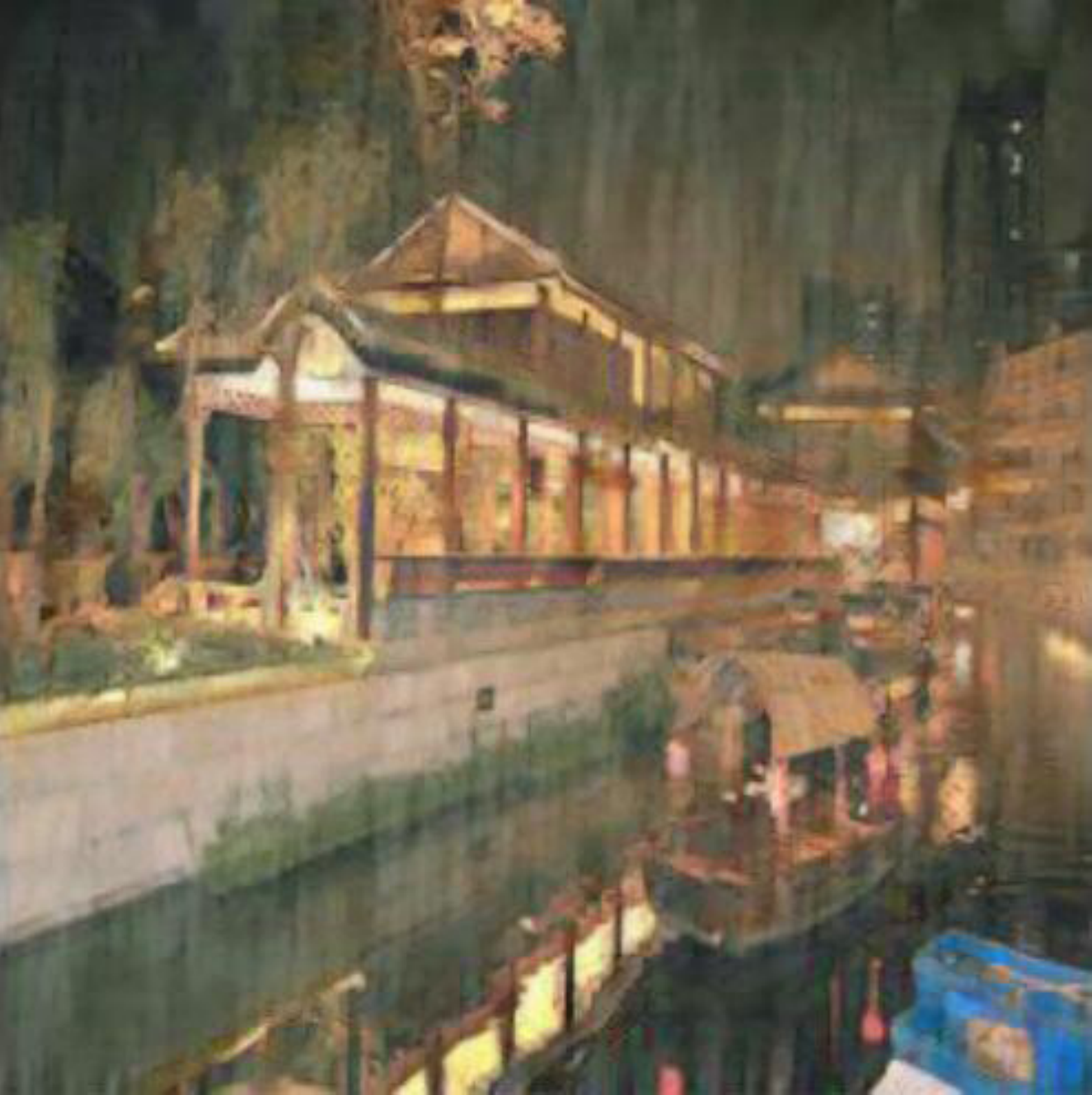}}
\end{minipage}
\hfill
\begin{minipage}{0.16\linewidth}
\centering{\includegraphics[width=1\linewidth]{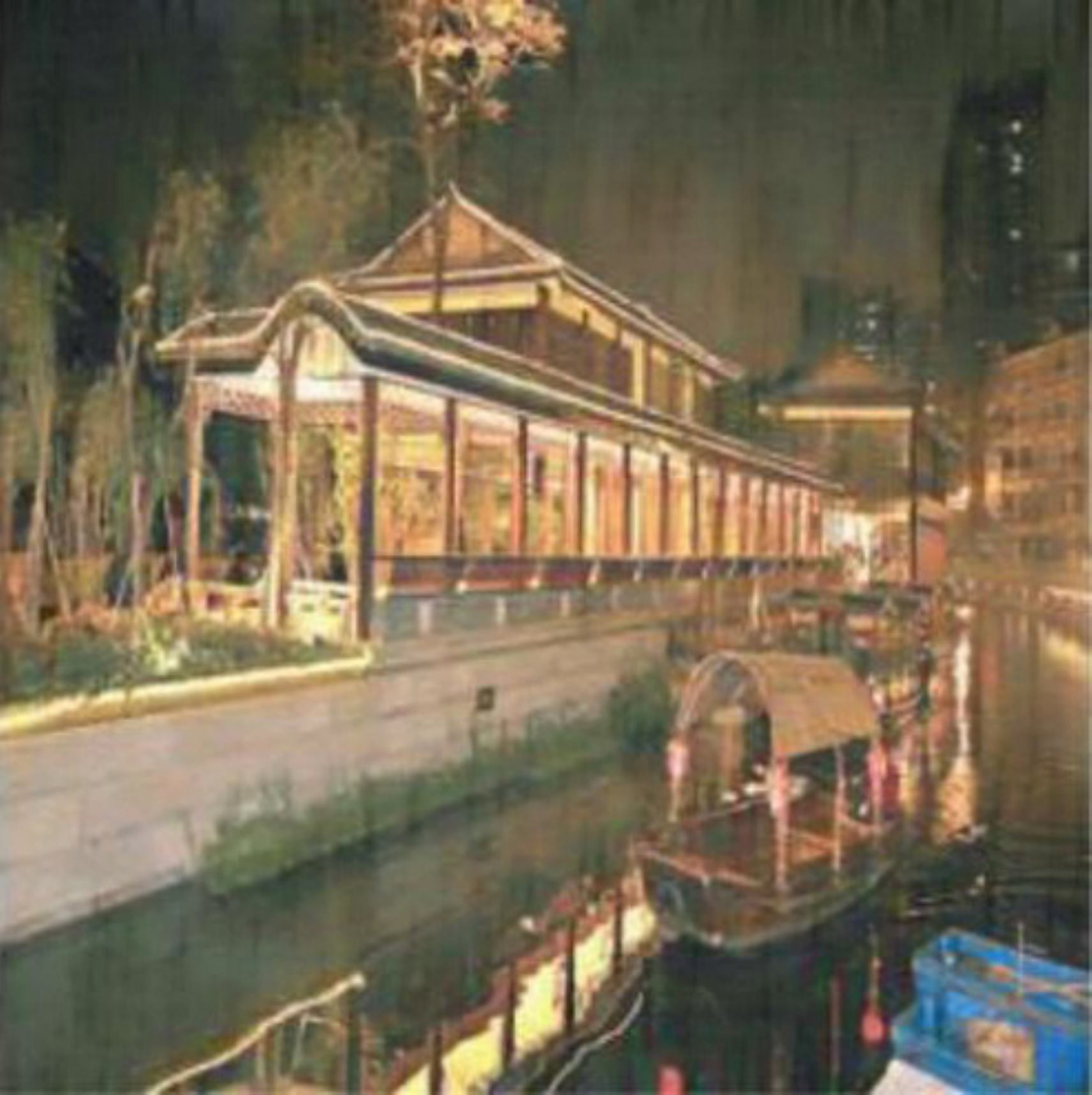}}
\end{minipage}
\vfill
\begin{minipage}{0.16\linewidth}
\centering{\includegraphics[width=1\linewidth]{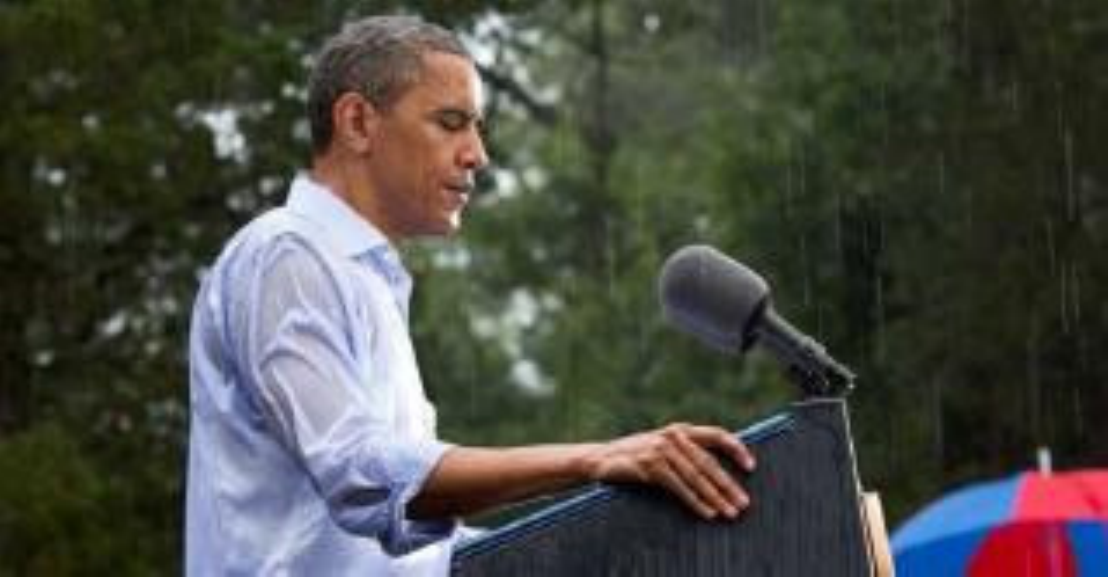}}
\centerline{Input}
\end{minipage}
\hfill
\begin{minipage}{0.16\linewidth}
\centering{\includegraphics[width=1\linewidth]{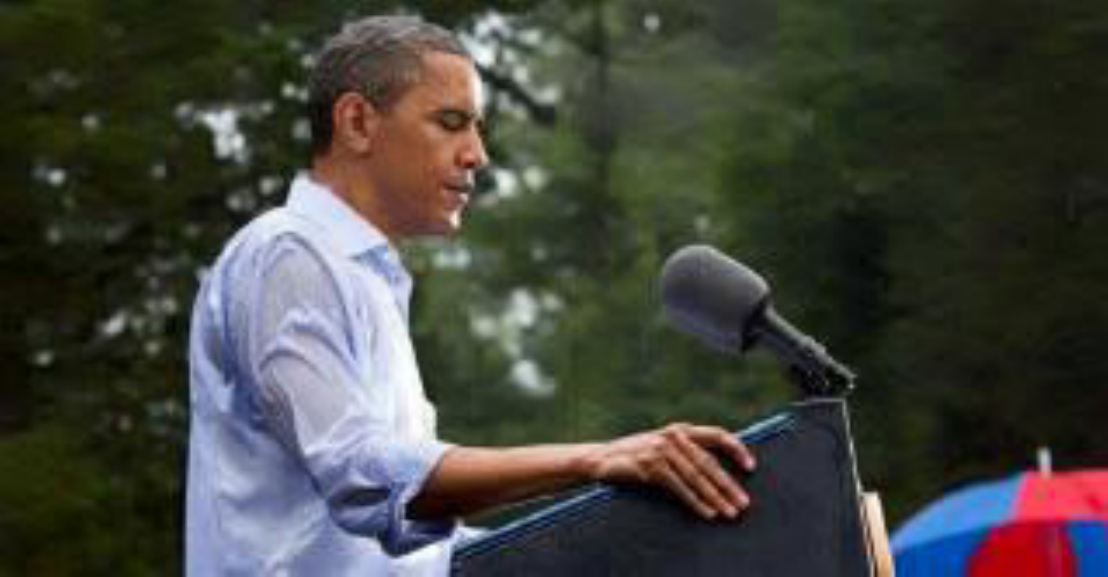}}
\centerline{\cite{Fu_2017_CVPR}}
\end{minipage}
\hfill
\begin{minipage}{0.16\linewidth}
\centering{\includegraphics[width=1\linewidth]{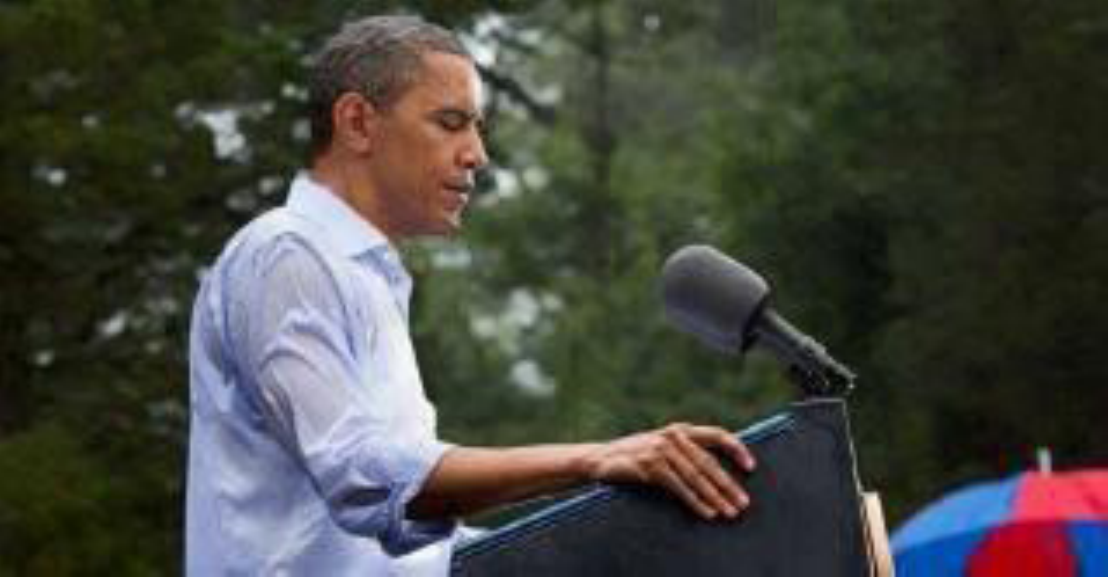}}
\centerline{\cite{Li_2018_ECCV}}
\end{minipage}
\hfill
\begin{minipage}{0.16\linewidth}
\centering{\includegraphics[width=1\linewidth]{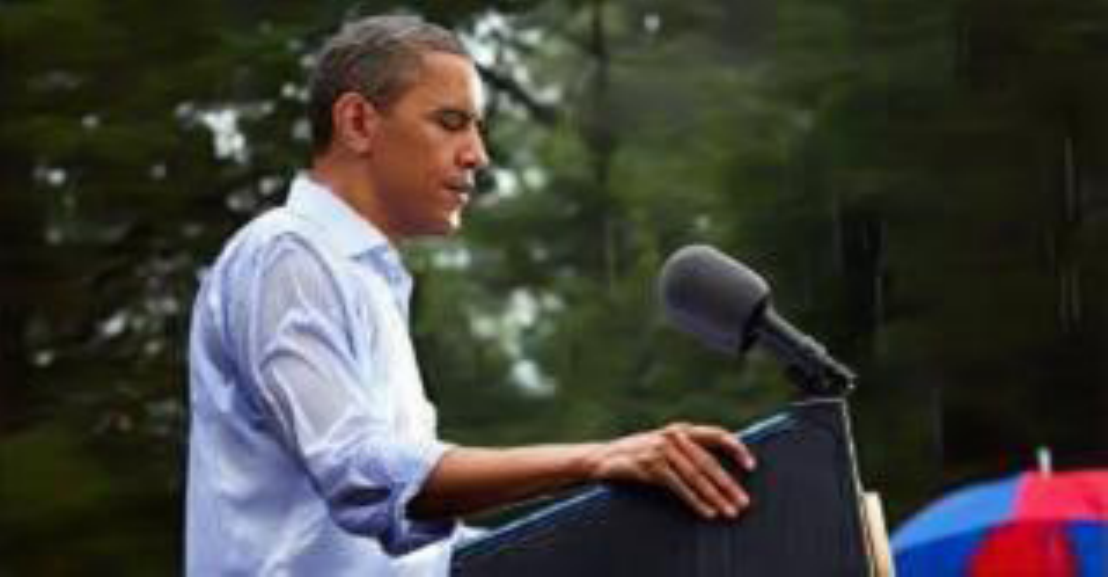}}
\centerline{\cite{Zhang_2018_CVPR}}
\end{minipage}
\hfill
\begin{minipage}{0.16\linewidth}
\centering{\includegraphics[width=1\linewidth]{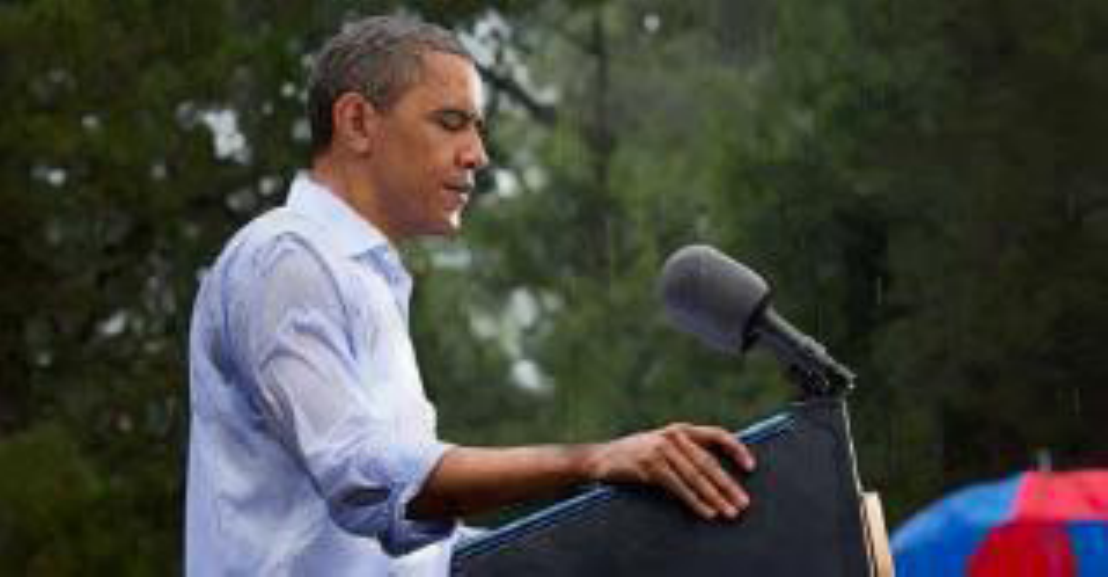}}
\centerline{\cite{Yang_2017_CVPR}}
\end{minipage}
\hfill
\begin{minipage}{0.16\linewidth}
\centering{\includegraphics[width=1\linewidth]{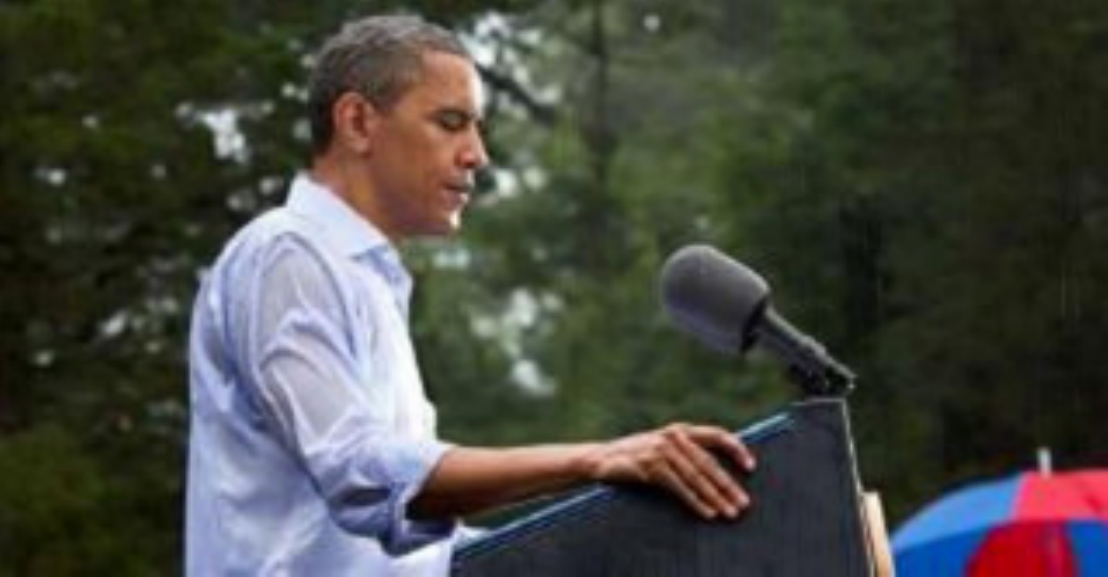}}
\centerline{Ours}
\end{minipage}
\end{center}
\caption{{This figure shows rain-removed results on some real-world rainy images.}}
\label{fig:practical_comparison}
\end{figure*}

\section{Experiments}

In order to assess the performance of our rain-removing method quantitatively, the commonly used PSNR and SSIM \cite{Wang_2004_TIP} are used as our metrics. For visual quality evaluation, we give the rain-removed results of some real-world and synthetic rainy images. Because the authors of \cite{Li_2018_MM} do not want to release their code so far, we utilize another four state-of-the-art works \cite{Yang_2017_CVPR,Zhang_2018_CVPR,Fu_2017_CVPR,Li_2018_ECCV} to make comparisons.

\subsection{Atmospheric Light and Transmission}

Before comparing with selected works, we show the atmospheric light and transmission estimated by our method first.
Figure \ref{fig:trans_atm} shows the results of two real-world rainy images. The atmospheric light estimated by the method in Section \ref{sec:atmospheric_light} tend to change relatively apparently for different rainy images, while the values estimated by $\mathcal{A}(\cdot)$ are much more stable. According to Eq. \eqref{eq:degradation_model}\eqref{eq:transmission}, when $d \to \infty$, $\mathbf{T(\mathbf{x})}=0$ and $\mathbf{I(\mathbf{x})}=\mathbf{A}$. Hence, $\mathbf{A}$ is the intensity of light before being scattered by medium, it should be a relatively stable and large value, hence our network $\mathcal{A}(\cdot)$ gives a better estimation to $\mathbf{A}$. We will also show later that the rain-removed results of using $\mathcal{A}(\cdot)$ are much better.

Our method obtains good de-raining results (Figure \ref{fig:trans_atm}(b)), and transmission gives us more information of rain. First, the shapes of rain streaks are more apparent, especially, some rain streaks merged in the background. Second, the influence of rain on background is also clearer, the pixels having high values attenuate much more light. Some other information, such as rain density and accumulation are also seen more apparently.


\subsection{Datasets}

In our work, we randomly select $6000$ training samples from the dataset in \cite{Zhang_2018_CVPR} and \cite{Li_2018_arxiv} respectively, so our training dataset includes $12000$ samples. One of our testing datasets Rain-I (has $400$ samples) is composed by randomly selecting $100$ samples from the testing datasets of \cite{Zhang_2018_CVPR,Li_2018_ECCV,Fu_2017_CVPR,Yang_2017_CVPR} respectively, which are selected works we to make comparisons. As the synthetic rainy images in \cite{Li_2018_arxiv} are different from others, our another testing dataset Rain-II ($400$ samples) is randomly selected from the testing dataset in this work. The real-world rainy images come from the previous works, and some of them are from Google.

\subsection{Quantitative Evaluation on Synthetic Datasets}

We show the PSNR/SSIM comparisons with the selected state-of-the-art rain-removing works in Table \ref{tab:psnr_ssim_comparison}. We can see that our method obtains higher PNSR/SSIM on both datasets. The PSNR is about $4$ dB higher than the best state-of-the-art \cite{Fu_2017_CVPR} on Rain-II. In this dataset, the synthetic rain streaks are wide and their edges are not clear and emerge into the background. Figure \ref{fig:synthetic_compare} are two synthetic rainy images that state-of-the-art cannot handle well. By comparison, our method produces better rain-removed results, nearly all rain streaks disappear. For the second one, though we remove the heavy rain streaks, some traces still remain in our results.
More results are in supplement.

\subsection{Qualitative Evaluation on Real-world Images}

In Figure \ref{fig:practical_comparison}, we show some results of selected and proposed methods on real-world images. When encountered with wide rain streaks, selected methods cannot produce desired results, some apparent rain streaks remain in the resulting images (\eg, the first one). For narrow rain streaks (\eg, third one), all the methods can acquire comparable results. For the rain streaks which have blur edges (\eg. the second one), the method \cite{Zhang_2018_CVPR} and our method obtain better results. The complexity comparisons are shown in Table \ref{tab:time_comparison}, we can see our work is the fastest one.

\subsection{Ablation Studies}

\begin{figure}[!t]
\begin{center}
\begin{minipage}{0.158\linewidth}
\centering{\includegraphics[width=1\linewidth]{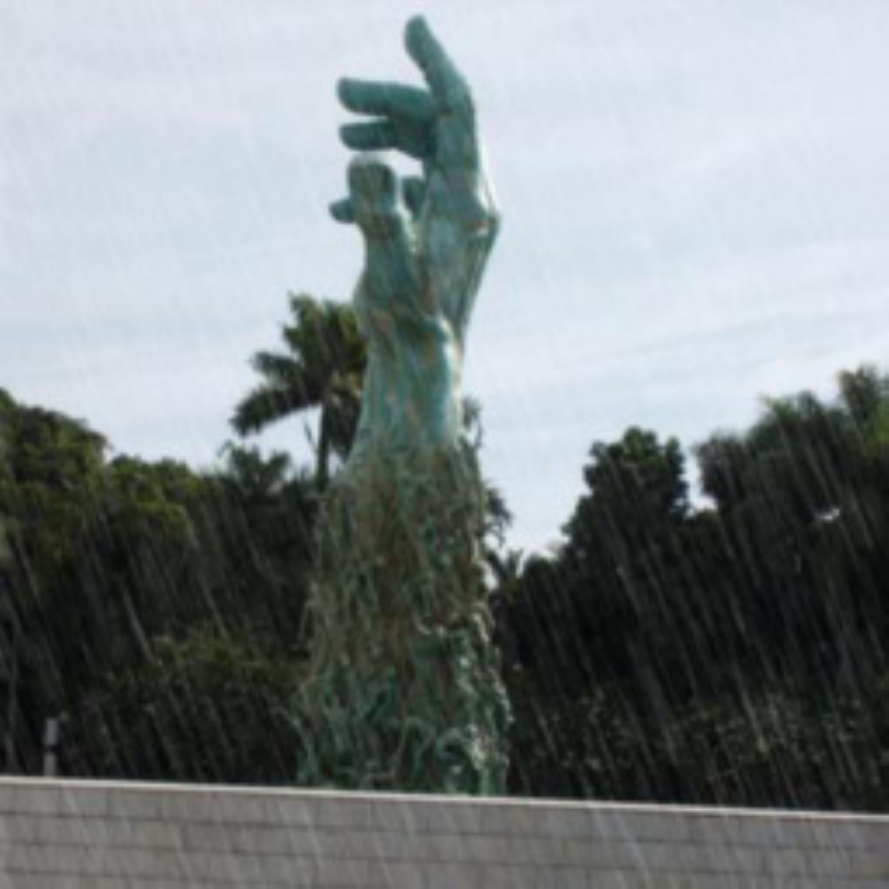}}
\end{minipage}
\hfill
\begin{minipage}{0.158\linewidth}
\centering{\includegraphics[width=1\linewidth]{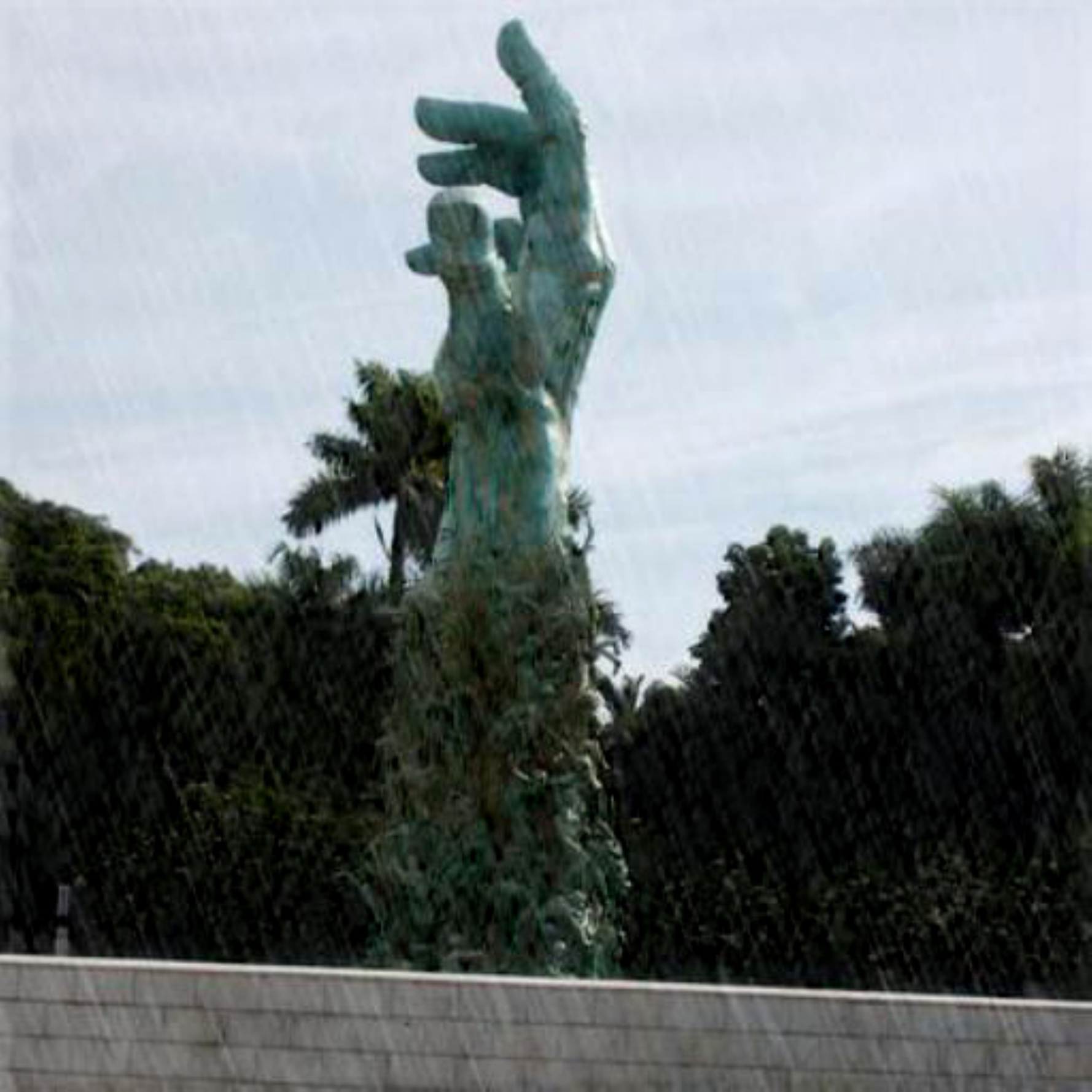}}
\end{minipage}
\hfill
\begin{minipage}{0.158\linewidth}
\centering{\includegraphics[width=1\linewidth]{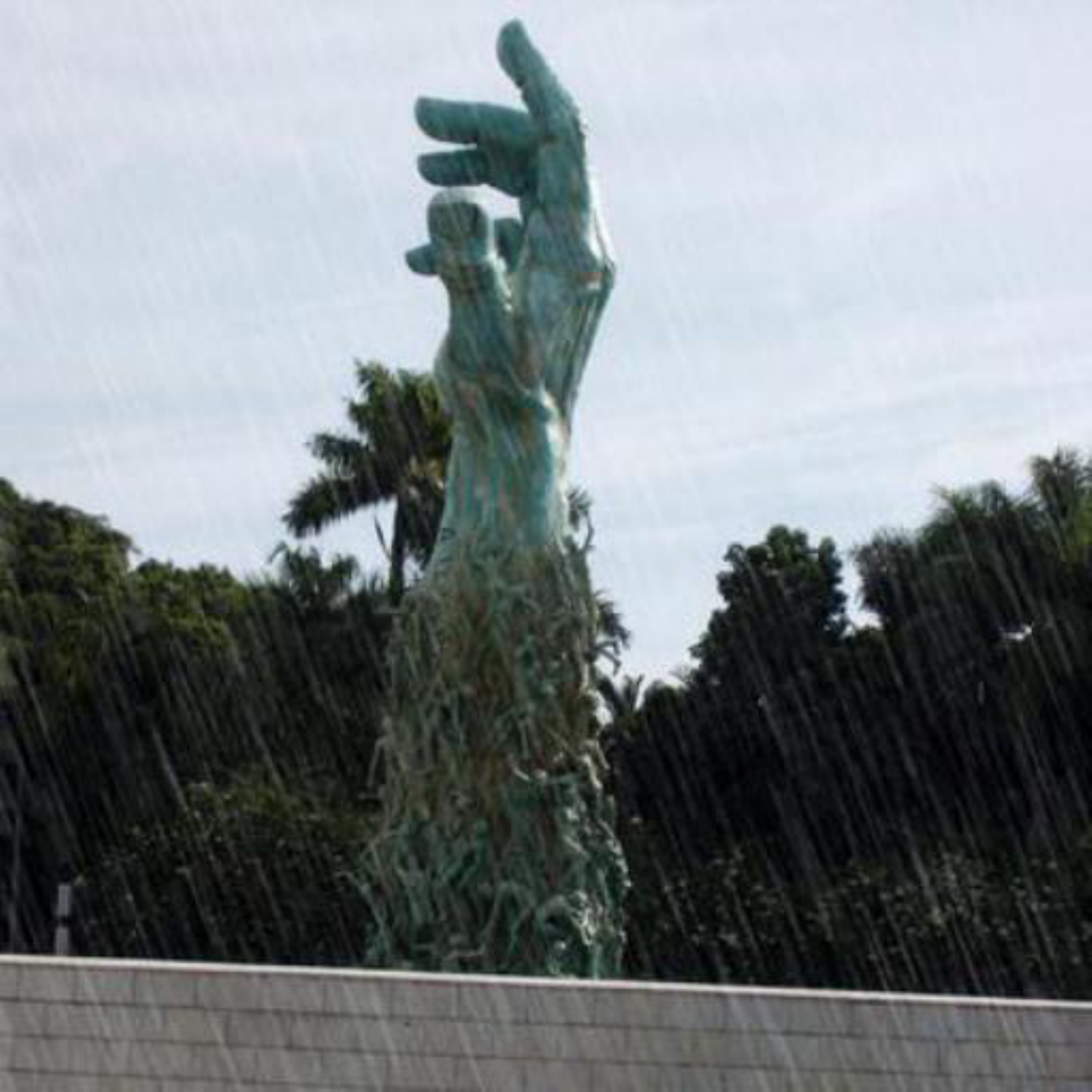}}
\end{minipage}
\hfill
\begin{minipage}{0.158\linewidth}
\centering{\includegraphics[width=1\linewidth]{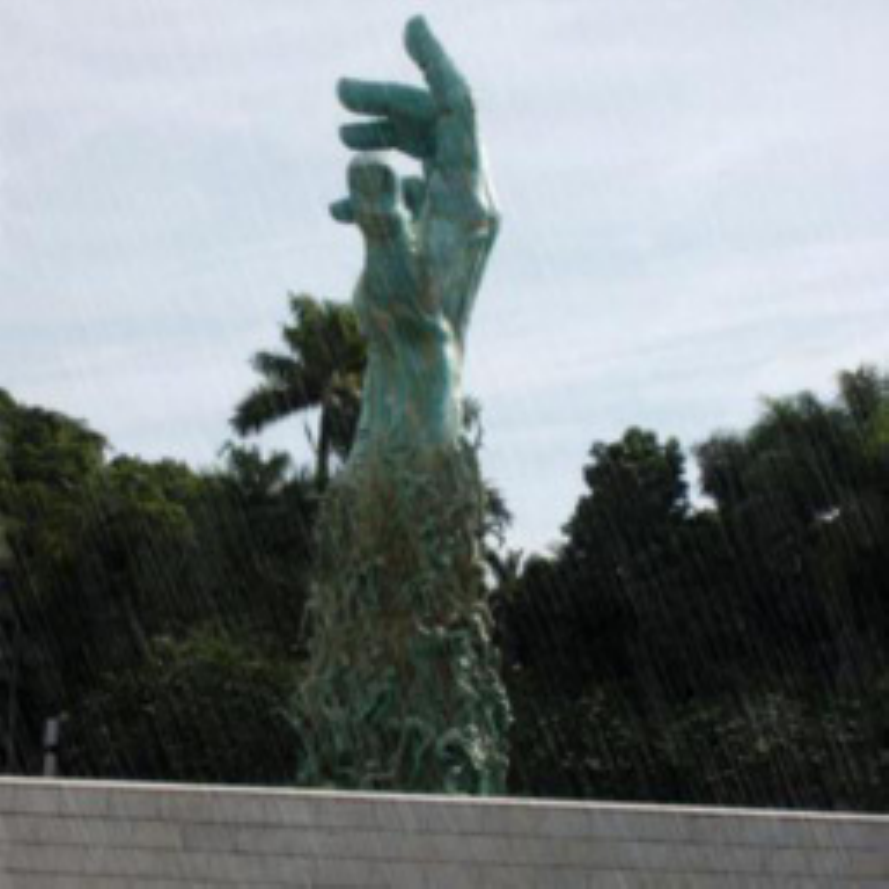}}
\end{minipage}
\hfill
\begin{minipage}{0.158\linewidth}
\centering{\includegraphics[width=1\linewidth]{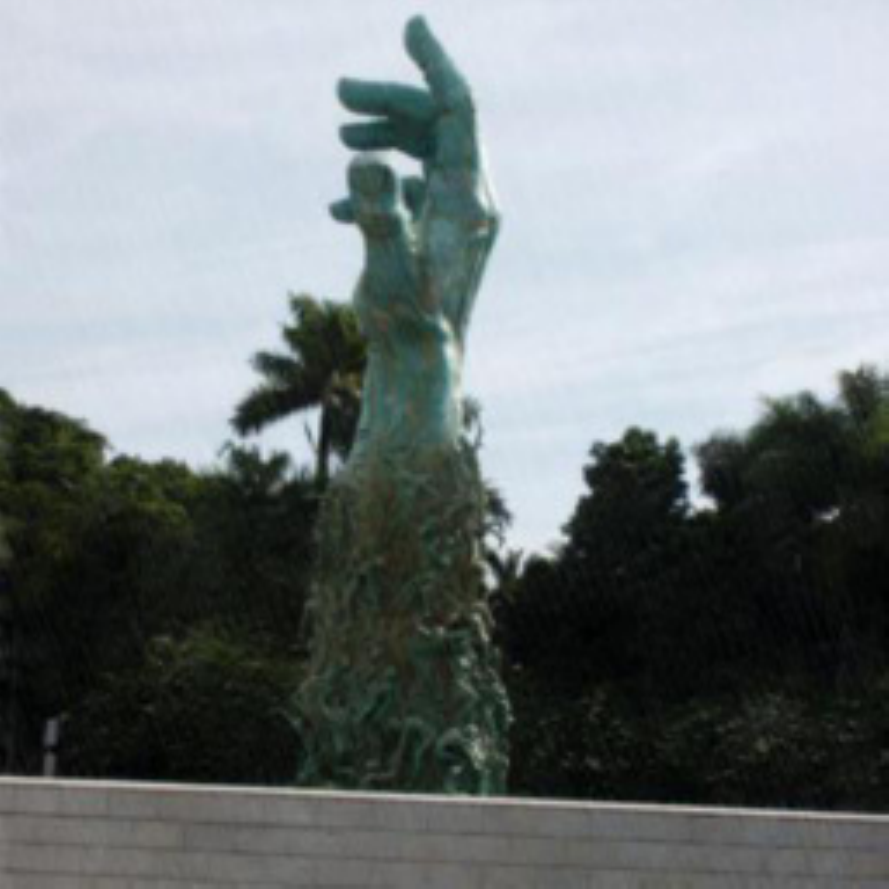}}
\end{minipage}
\hfill
\begin{minipage}{0.158\linewidth}
\centering{\includegraphics[width=1\linewidth]{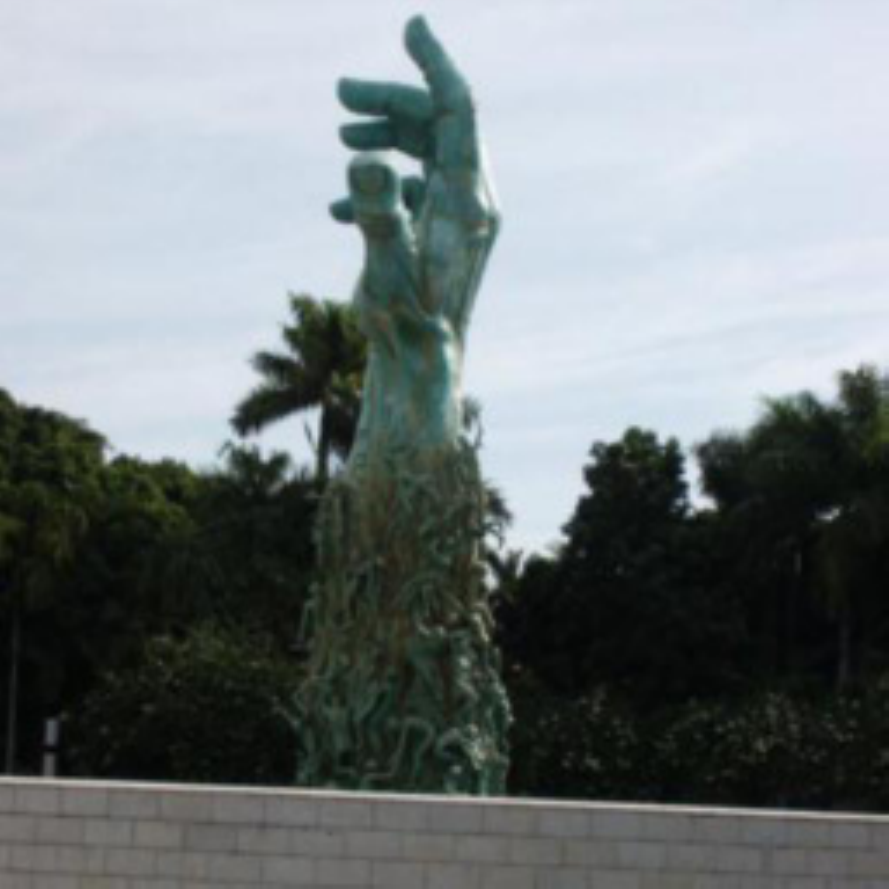}}
\end{minipage}
\vfill
\begin{minipage}{0.19\linewidth}
\centering{\includegraphics[width=1\linewidth]{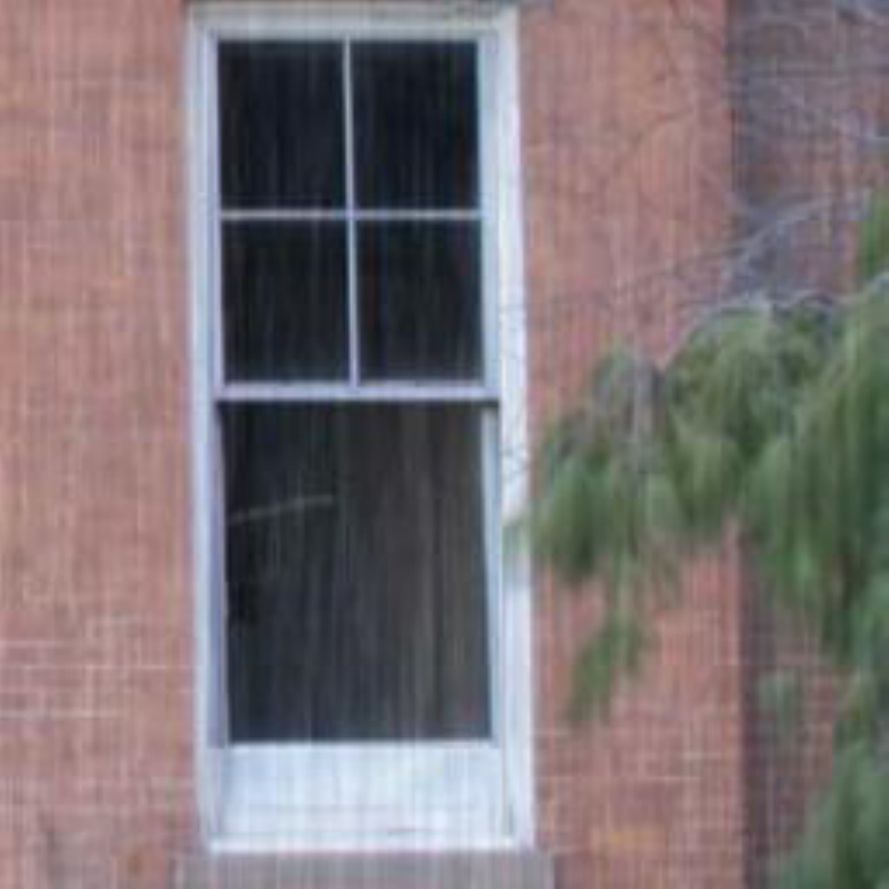}}
\end{minipage}
\hfill
\begin{minipage}{0.19\linewidth}
\centering{\includegraphics[width=1\linewidth]{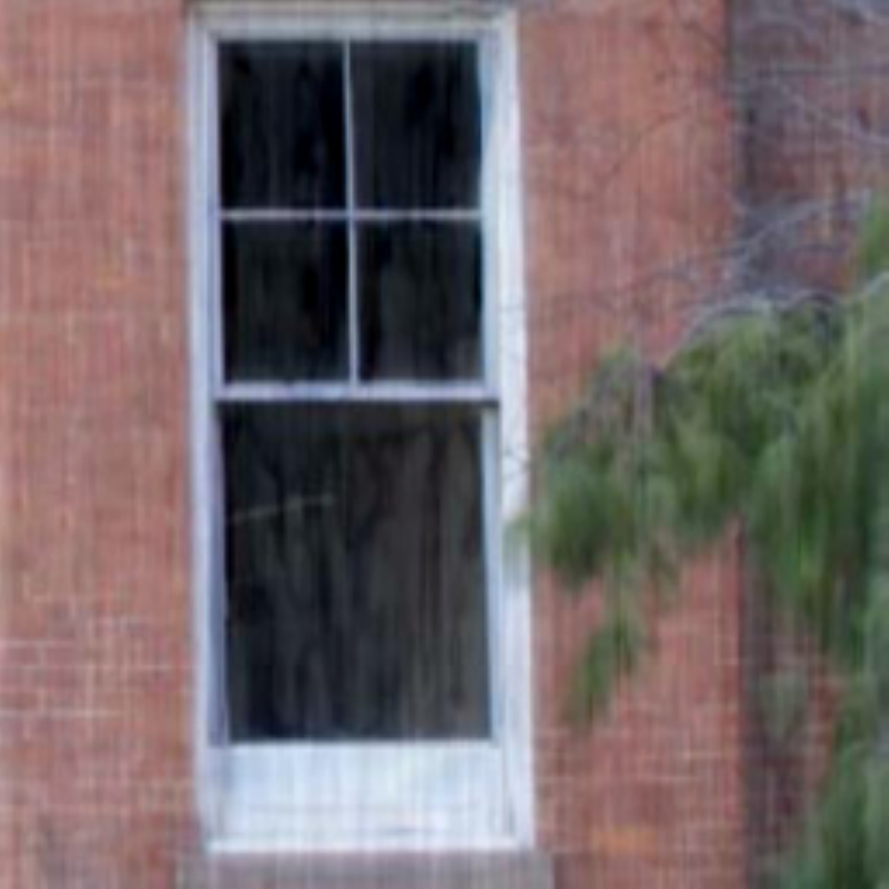}}
\end{minipage}
\hfill
\begin{minipage}{0.19\linewidth}
\centering{\includegraphics[width=1\linewidth]{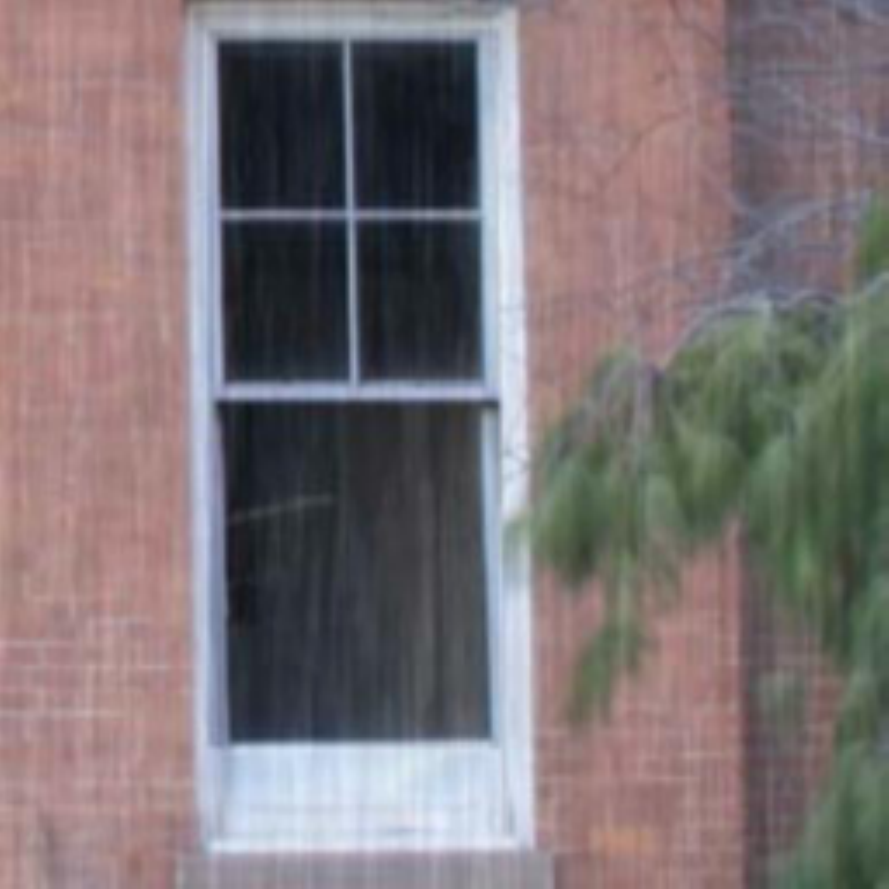}}
\end{minipage}
\hfill
\begin{minipage}{0.19\linewidth}
\centering{\includegraphics[width=1\linewidth]{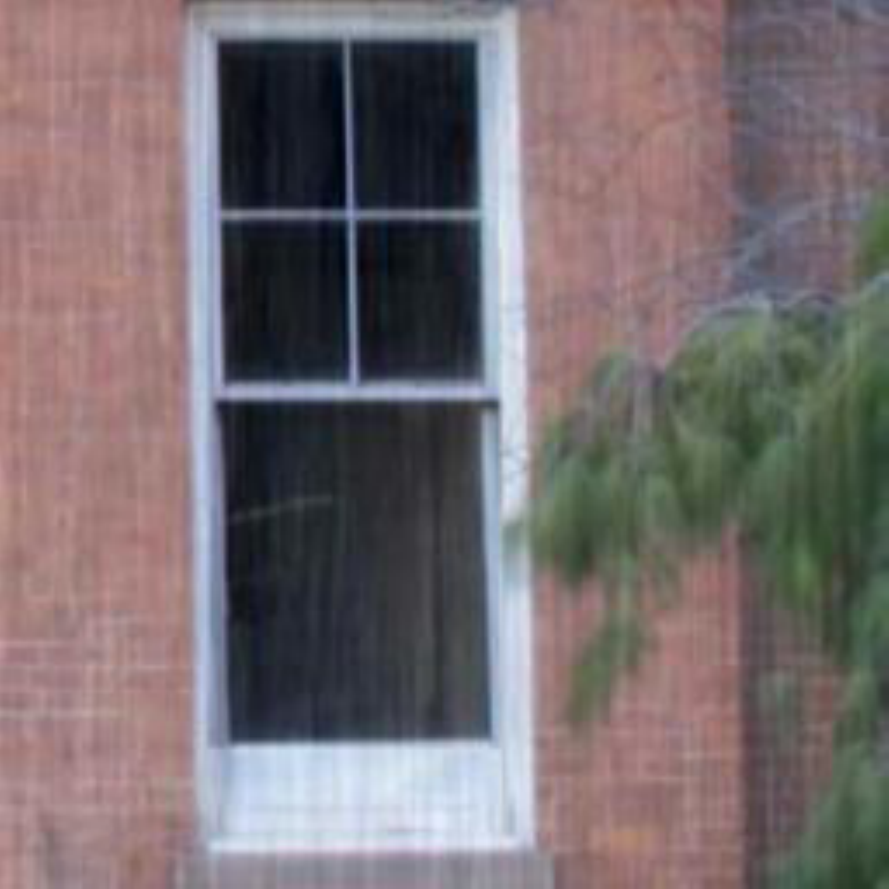}}
\end{minipage}
\hfill
\begin{minipage}{0.19\linewidth}
\centering{\includegraphics[width=1\linewidth]{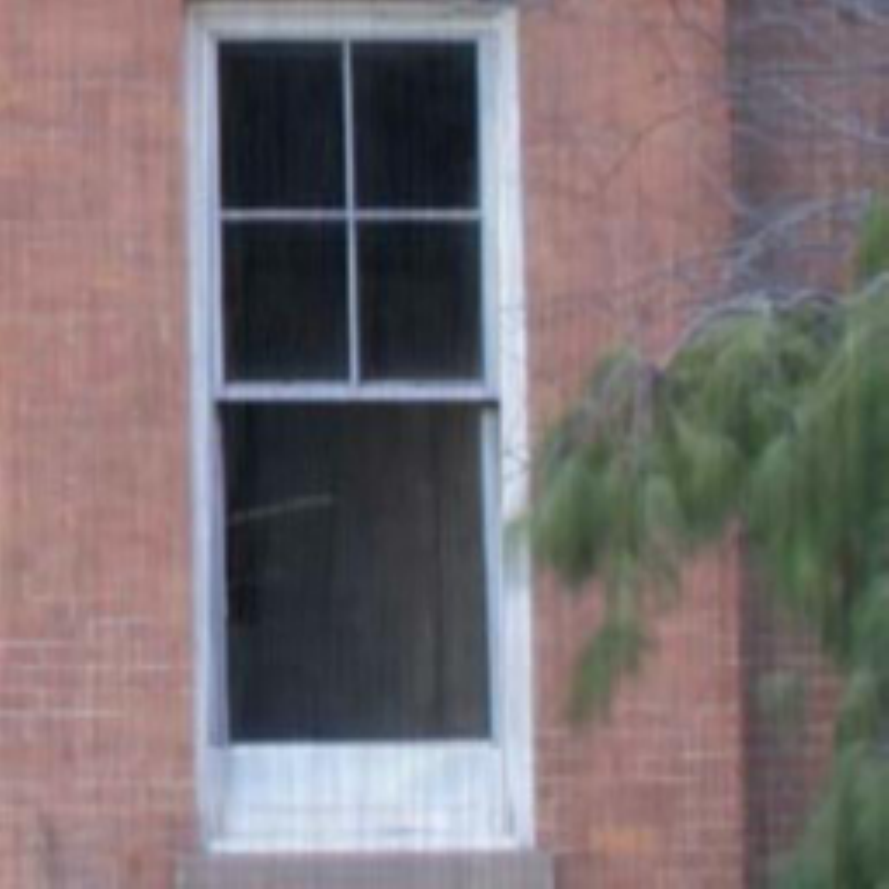}}
\end{minipage}
\end{center}
\caption{The first line is a synthetic rainy image and the second is a real-world one. From left to right, the images are input, results by DEMO-NetA1, DEMO-NetA2, DEMO-NetA3 and DEMO-Net. The last image in the first line is ground truth.}
\label{fig:ablation_syn_real}
\end{figure}

To verify the performances of our whole network further, we do some ablation experiments. Our network has three different ablation variants. One is removing $\mathcal{A}(\cdot)$, and using atmospheric light estimated by the method in Section \ref{sec:atmospheric_light} to train the transmission network $\mathcal{T}(\cdot)$, we call it DEMO-NetA1 to simplify our expression.
The second one is that $\mathcal{A}(\cdot)$ and $\mathcal{T}(\cdot)$ are trained jointly, but the pre-trained parameters for $\mathcal{A}(\cdot)$ is not utilized, which is called DEMO-NetA2.
In the last one, we use pre-trained $\mathcal{A}(\cdot)$ to estimate atmospheric light, but its parameters will not be updated during the training for $\mathcal{T}(\cdot)$, we call it DEMO-NetA3. In Figure \ref{fig:ablation_syn_real} we show the ablation study results. The objective indexes are shown in Table \ref{tab:ablation_comparison}. We can see that DEMO-NetA1 has lowest performance, which may be related to the relatively large variance of estimated $\mathbf{A}$ for different images by the method in Section \ref{sec:atmospheric_light}. This situation has good convergence during the training, but its generalization is low. With the introduce of $\mathcal{A}(\cdot)$, the performance is enhanced. DEMO-NetA3 is better than DEMO-NetA2, which proves that our estimated $\mathbf{A}$ in Section \ref{sec:atmospheric_light} is useful. Our DEMO-Net possesses the apparent best performances, we use the pre-trained parameters for $\mathcal{A}(\cdot)$, then fine-tuned them during the training of $\mathcal{T}(\cdot)$, which is equal to adding a prior of the global atmospheric light to the training of the whole network.

\begin{table}[]
\centering
\caption{PSNR and SSIM of our ablation studies.}
\begin{tabular}{c|cc|cc}
\hline
Variants & \multicolumn{2}{c|}{Rain-I} & \multicolumn{2}{c}{Rain-II} \\  \hline
Metric &    PSNR       &    SSIM       &    PSNR       &    SSIM       \\ \hline
DEMO-NetA1 &   27.15        &   0.772        &    25.48       &   0.793        \\
DEMO-NetA2 &   27.49        &    0.806       &    28.57       &     0.844      \\
DEMO-NetA3 &   28.13        &    0.843       &    29.07       &    0.874       \\ \hline
DEMO-Net &    \textbf{31.35}       &    \textbf{0.898}       &    \textbf{34.41}       &    \textbf{ 0.936}      \\ \hline
\end{tabular}
\label{tab:ablation_comparison}
\end{table}

\begin{figure}[!t]
\begin{center}
\begin{minipage}{0.32\linewidth}
\centering{\includegraphics[width=1\linewidth]{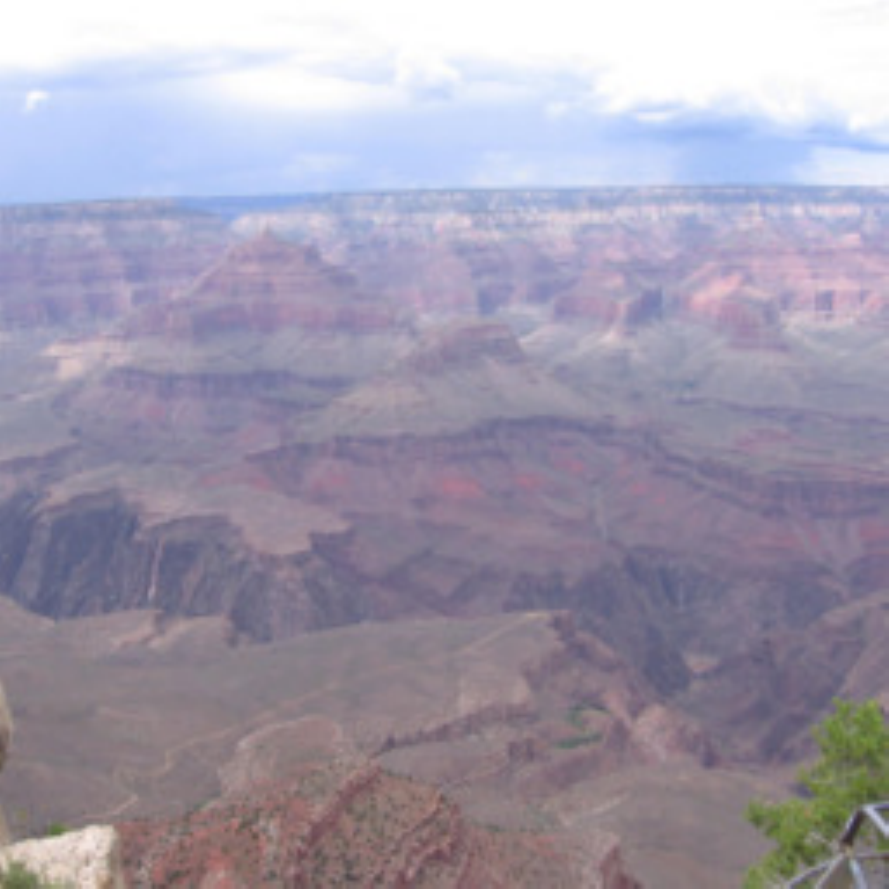}}
\end{minipage}
\hfill
\begin{minipage}{0.32\linewidth}
\centering{\includegraphics[width=1\linewidth]{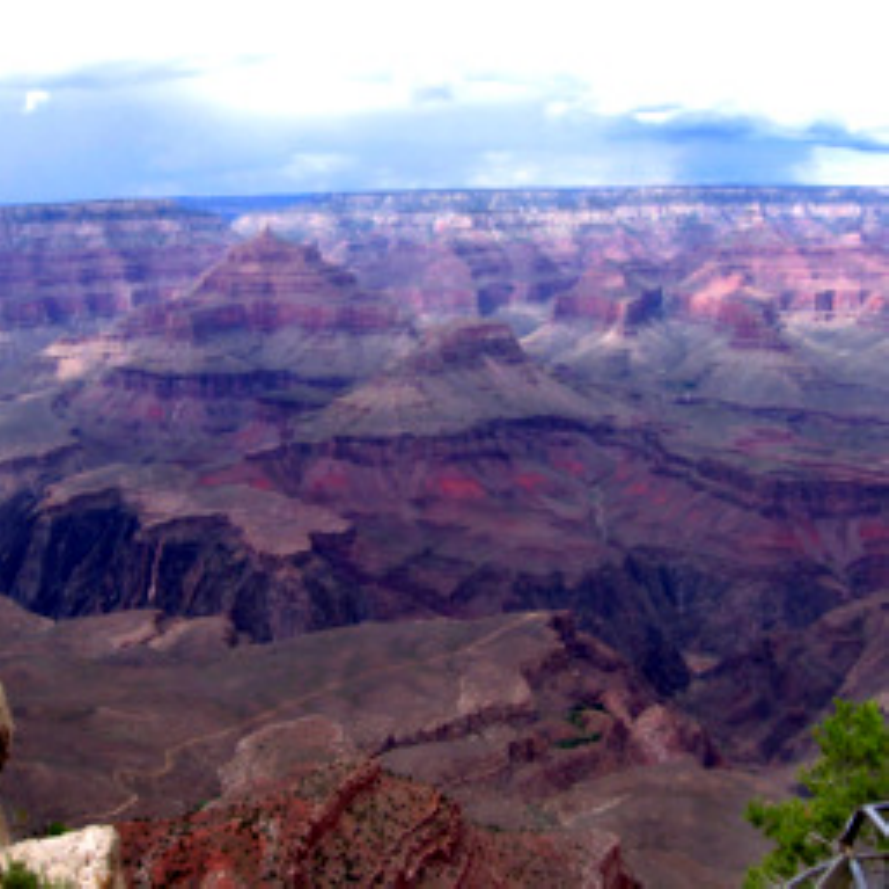}}
\end{minipage}
\hfill
\begin{minipage}{0.32\linewidth}
\centering{\includegraphics[width=1\linewidth]{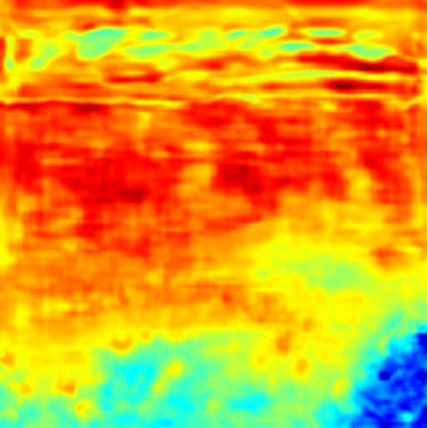}}
\end{minipage}
\vfill
\begin{minipage}{0.32\linewidth}
\centering{\includegraphics[width=1\linewidth]{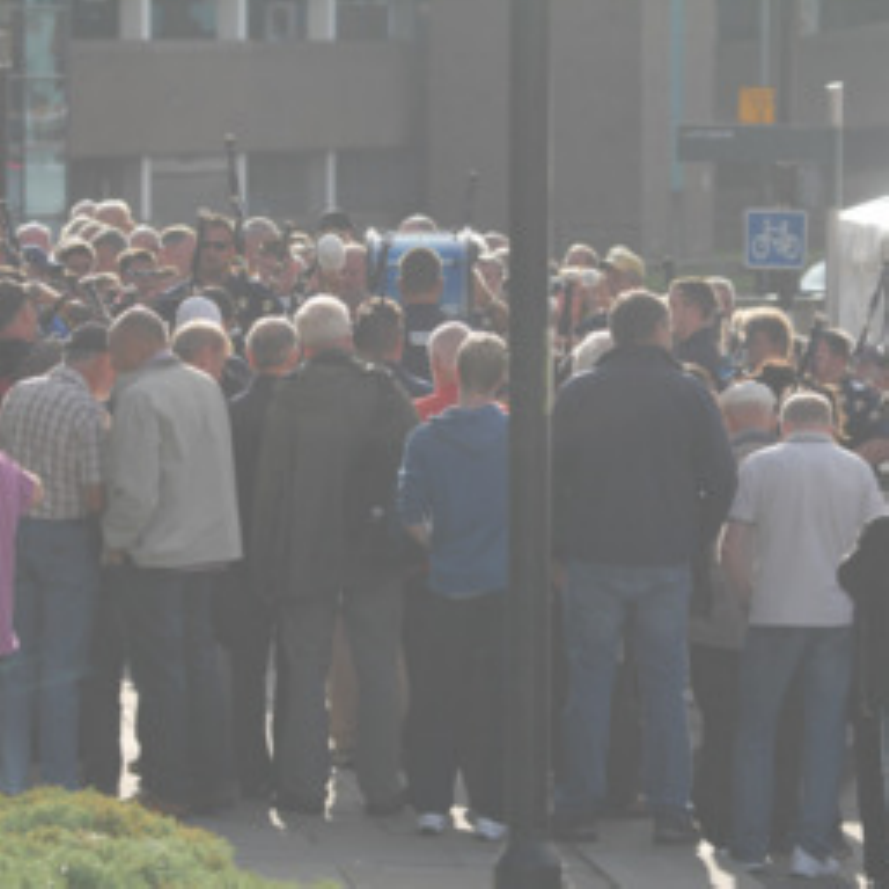}}
\centerline{Hazy images}
\end{minipage}
\hfill
\begin{minipage}{0.32\linewidth}
\centering{\includegraphics[width=1\linewidth]{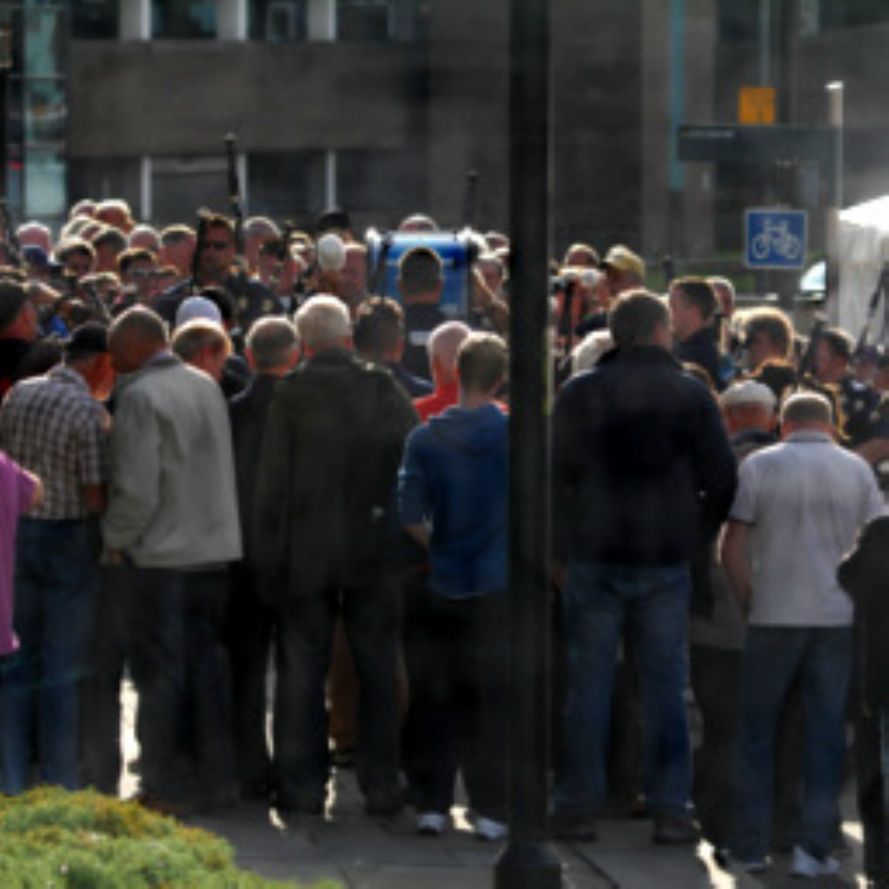}}
\centerline{Dehazing results}
\end{minipage}
\hfill
\begin{minipage}{0.32\linewidth}
\centering{\includegraphics[width=1\linewidth]{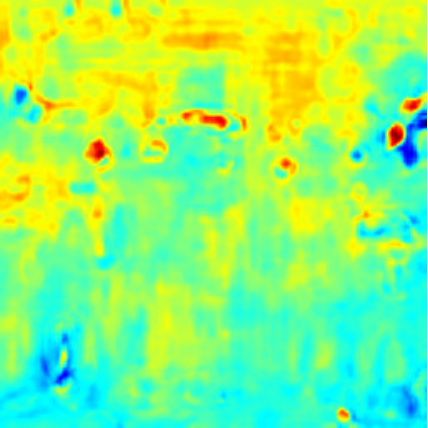}}
\centerline{Transmission}
\end{minipage}
\end{center}
\caption{{This figure shows some dehazing results and obtained transmission by our method for hazy images. Our network is trained on a datase with just $5000$ training samples.}}
\label{fig:dehaze}
\end{figure}

\subsection{Other Potentials of Our Network}

Our work still has some potentials to handle images obtained under other weather conditions. In Figure \ref{fig:dehaze} we simply show some dehazing results by our DEMO-Net. Besides, if the initial $\mathbf{A}$ for snow can be obtained, maybe our method can also work on snowy images.

\section{Conclusions}

In this paper, we utilize an image degradation model which is derived from atmospheric scattering principle to model the formation of rainy images. In order to remove rain effect and obtain the transmission which contain more information of rain, we proposed a method based on the scattering principle to estimate the global atmospheric light initially. To estimate a better global atmospheric light and enhance the de-raining performance, we designed a special triangle-shaped network to learn global atmospheric light for rainy images supervised by the initial estimation. Then a simple but novel network which boots information pass was used to learn the transmission, during this process, the triangle-shaped network is fine-tuned further. Results on synthetic and real-world rainy images show that our method outperforms the selected state-of-the-art, and obtains a transmission which contain more information in rainy scene.


\end{document}